\pdfoutput=1

\documentclass[11pt]{article}

\usepackage{acl2022}
\aclfinalcopy
\usepackage{times}
\usepackage{latexsym}

\usepackage[export]{adjustbox}
\usepackage{adjustbox}
\usepackage{booktabs}
\usepackage{multirow}

\usepackage{microtype}
\usepackage{amsmath}
\usepackage{hyperref}
\usepackage{cleveref}

\crefname{section}{§}{§§}
\Crefname{section}{§}{§§}
\usepackage{arydshln}
\usepackage[noend]{algpseudocode}
\usepackage{color,soul}
\usepackage{xspace}
\usepackage{mathtools}
\usepackage{amssymb}
\usepackage{footnote}
\usepackage{tablefootnote}
\usepackage{graphicx}
\usepackage{color, colortbl}
\usepackage{xstring}
\usepackage{comment}
\usepackage{adjustbox}
\usepackage{float}
\restylefloat{table}
\usepackage{graphicx}
\usepackage{array,booktabs,makecell}
\usepackage{multirow}
\usepackage{graphicx}
\usepackage{appendix}
\usepackage[show]{boxnotes}
\newcommand{\eat}[1]{} 
 
\usepackage{arabtex}

\iftrue 

\fi
\usepackage{array}
\newcolumntype{H}{>{\setbox0=\hbox\bgroup}c<{\egroup}@{}}
\usepackage{xcolor}
\usepackage{color, colortbl}
\usepackage{utf8}
\usepackage[utf8]{inputenc}
\usepackage{cancel}
\usepackage{ulem,xpatch}
\usepackage[T1]{fontenc}

\usepackage{microtype}

\newcommand\blfootnote[1]{%
  \begingroup
  \renewcommand\thefootnote{}\footnote{#1}%
  \addtocounter{footnote}{-1}%
  \endgroup
}

\title{AraT5: Text-to-Text Transformers for Arabic Language Generation}

\author{El Moatez Billah Nagoudi$^{\star}$ ~~~ AbdelRahim Elmadany$^{\star}$ ~~~Muhammad Abdul-Mageed$^{\star}$ \\
\normalsize Deep Learning and Natural Language Processing Group  \\
  \normalsize The University of British Columbia\\
      
  \texttt{ \small \{moatez.nagoudi,a.elmadany,muhammad.mageed\}@ubc.ca} }

\begin{document}
\setcode{utf8}
\maketitle




\section*{~~~~~~~~~~~~~~~~~~~~~~~~~~~~~Abstract}

  Transfer learning with a unified Transformer framework (T5) that converts all language problems into a text-to-text format was recently proposed as a simple and effective transfer learning approach. Although a multilingual version of the T5 model (mT5) was also introduced, it is not clear how well it can fare on non-English tasks involving \textit{diverse} data. To investigate this question, we apply mT5 on a language with a wide variety of dialects--Arabic. For evaluation, we introduce a novel benchmark for \textbf{AR}abic language \textbf{GEN}eration (ARGEN), covering \textit{seven} important tasks. For model comparison, we pre-train three powerful Arabic T5-style models and evaluate them on ARGEN. Although pre-trained with $\sim49\%$ less data, our new models perform significantly better than mT5 on \textit{all} ARGEN tasks (in $52$ out of $59$ test sets) and set several new SOTAs. Our models also establish new SOTA on the recently-proposed, large Arabic language understanding evaluation benchmark ARLUE~\cite{abdul2020arbert}. Our new models are publicly available. We also link to ARGEN datasets through our repository.\footnote{\href{https://github.com/UBC-NLP/araT5}{https://github.com/UBC-NLP/araT5}}  \\~\blfootnote{ $^{\star}$ All authors contributed equally.}  \\
  
\label{sec:abs}




\section{Introduction}\label{sec:intro}

Due to their remarkable ability to transfer knowledge from unlabeled data to downstream tasks, pre-trained Transformer-based language models have emerged as important components of modern natural language processing (NLP) systems. In particular, the unified framework that converts all text-based language problems into a text-to-text format presented through the T5 model~\cite{raffel2019exploring} is attractive. In addition to its simplicity, this approach is effective since it allows knowledge transfer from high-resource to low-resource tasks without the need for changing model architecture. Unlike models such as BERT~\cite{devlin2019bert}, which are based on encoders only, the T5 model is an encoder-decoder that can naturally be employed for natural language generation. 
\begin{figure}[t]
  \centering
  \includegraphics[width=\linewidth]{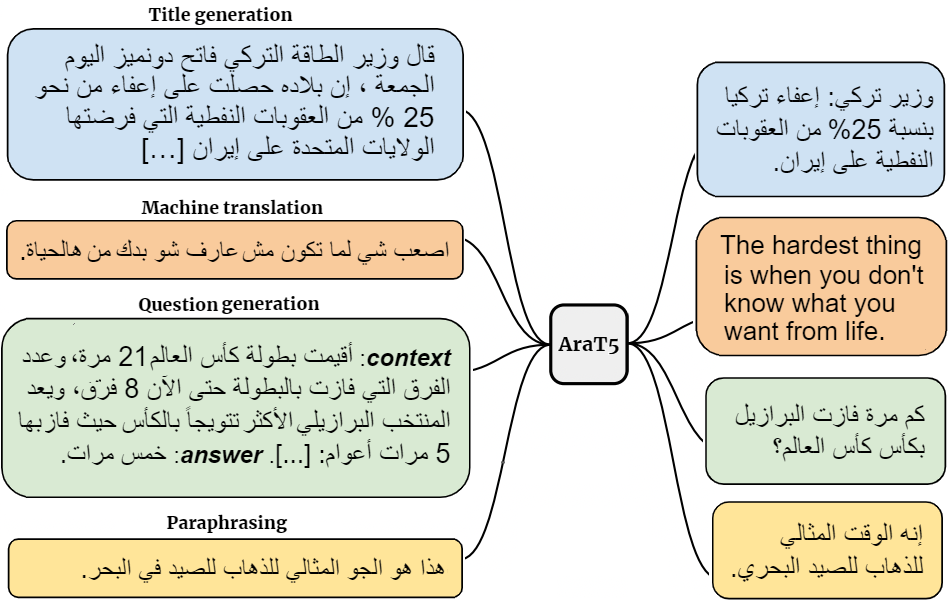}
\caption{\small Our AraT5 encoder-decoder model and prompt samples from four investigated tasks, namely: title generation, machine translation, question generation, and paraphrasing.}
\label{fig:AraT_overview} 
\end{figure}
Although the T5 model, originally pre-trained for English, was recently extended to the multilingual setting as mT5~\cite{xue2020mt5}, it is not clear how suited it is to individual languages (and varieties of these languages). In addition, systematic issues have been discovered in multilingual corpora on which language models have been trained~\cite{caswell2021quality}. In absence of comparisons with monolingual pre-trained language models that serve different non-English contexts, it remains unknown how multilingual models really fare against language-specific models. 

In this work, we offer the first comparison of the mT5 model to similar encoder-decoder models dedicated to Arabic. We choose Arabic as our context due to its large set of diverse varieties as well as its wide use on social media. Our work aims at uncovering the extent to which mT5 can serve Arabic's different varieties. Our work also meets an existing need 
for pre-trained Transformer-based sequence-to-sequence models. In other words, while several BERT-based models have been pre-trained for Arabic~\cite{antoun2020arabert, abdul2020arbert, CAMeLBERT2021}, no such attempts have been made to create sequence-to-sequence models that we know of. Another motivation for our work is absence of an evaluation benchmark for Arabic language generation tasks. Apart from machine translation where researchers are starting to propose benchmarks such as AraBench~\cite{sajjad2020arabench}, there are no benchmarks that can be used to methodically measure Arabic natural language generation performance. 

Our main contributions are as follows: \textbf{(1)} We introduce three powerful variants of the text-to-text transformer (T5) model dedicated to Modern Standard Arabic (MSA) and a diverse set of Arabic dialects. We include in our vocabulary $11$ languages other than Arabic (e.g., English, French, German, Russian), which also allows us to evaluate our models under zero-shot pre-training conditions involving these languages. \textbf{(2)} We propose a novel unified benchmark for \textbf{AR}abic natural language \textbf{GE}eneration  (\textbf{ARGEN}) composed of \textit{seven} tasks: machine translation, code-switched text translation, summarization, news title generation, question generation, paraphrasing, and transliteration. ARGEN is collected  from a total of $19$ datasets, including  $9$ \textit{new  datasets} proposed in this work. \textbf{(3)}~To show the utility of our new models, we evaluate them on ARGEN under both \textit{full} and \textit{zero-shot} pre-training conditions. Our models set new SOTA on the majority of datasets in \textit{all} seven tasks. \textbf{(4)} Although the main focus of our work is language \textit{generation}, we also show the effectiveness of our models on Arabic language \textit{understanding} by fine-tuning our new models on a large, recently proposed Arabic language understanding benchmark. Again, our models establish new SOTA on the majority of language understanding tasks.






The rest of the paper is organized as follows: Section~\ref{sec:OUR_models} describes our Arabic pre-tained models. In Section~\ref{sec:AraLG}, we introduce ARGEN, our new natural language generation benchmark. We evaluate our models on ARGEN in Section~\ref{sec:EVAL}. Section~\ref{sec:ana_disc} is an analysis and discussion of our results. In Section~\ref{sec:RW}, we provide an overview of related work. We conclude in Section~\ref{sec:conclusion}. We now introduce our new pre-trained models.


\label{sec:intro}
\section{Our Models}\label{sec:OUR_models}

\subsection{Pre-Training Data}\label{subsec:tr_data}

\noindent\textbf{MSA Data.} We use $70$GB of MSA text ($7.1$B tokens) from the following sources: AraNews~\cite{nagoudi2020machine}, El-Khair~\newcite{elkhair-2016}, Gigaword,\footnote{ \href{https://catalog.ldc.upenn.edu/LDC2009T30}{https://catalog.ldc.upenn.edu/LDC2009T30}.}, OSCAR~\cite{suarez2019asynchronous}, OSIAN~\cite{zeroual2019osian},  Wikipedia Arabic, and Hindawi Books.\footnote{\href{https://www.hindawi.org/books/}{https://www.hindawi.org/books}.}


\noindent\textbf{Twitter Data.} We randomly sample $1.5$B Arabic tweets ($178$GB) from a large in-house dataset of $\sim10$B tweets. We use string matching to only include tweets with at least $3$ Arabic words, regardless whether the tweet has non-Arabic string or not.  

Our combined MSA and Twitter data make up $29$B tokens, and hence is $\sim49\%$ less than Arabic tokens on which mT5 is pre-trained ($57$B Arabic tokens). More information about our pre-training data is in Table~\ref{tab:art5_data}. 

\begin{figure}[h]

  \includegraphics[scale=0.381, left]{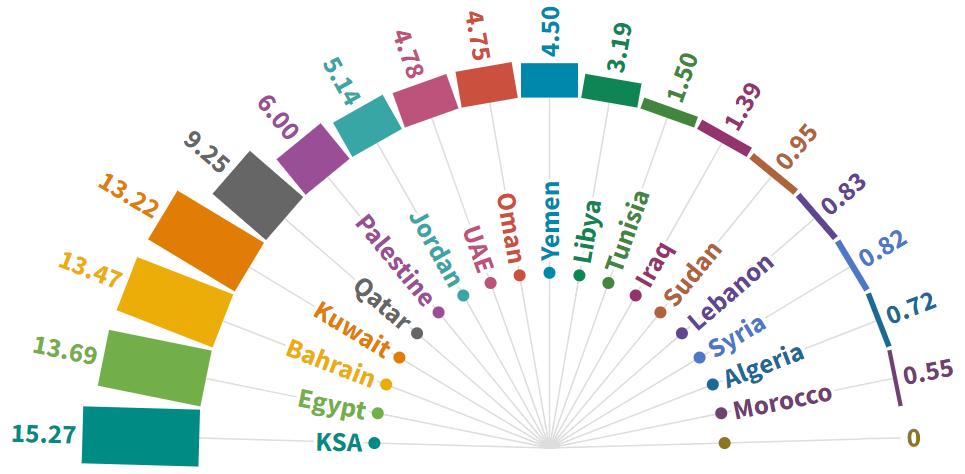}

\caption{ Country-level distribution in the dialectal portion of our data.}
\label{fig:geo-count_m}
\end{figure}

\noindent\textbf{MSA Vs. Dialect Distribution.} In order to analyze MSA-dialect distribution in our Twitter data, we run the binary (MSA-dialect) classifier introduced in~\citet{magee-2020-toward} on a random sample  of $100$M tweets. We find the data to involve $28.39$\% predicted dialect tweets and $71.61$\% predicted MSA. We also acquire country-level dialect labels using an in-house strong classifier on the dialectal portion of the data (i.e., $\sim28.39$  millions tweets),  finding dialectal tweets to be truly \textit{geographically diverse} as shown in Figure~\ref{fig:geo-count_m}.





\noindent{\textbf{Naturally-Occurring Code-Switching.}} Using $1$M random tweets from our data, we perform an analysis of code-switching. For this, we employ simple string matching to identify Arabic and run the CLD3 language ID tool\footnote{\href{https://github.com/google/cld3}{https://github.com/google/cld3}} on the non-Arabic string sequences. We find the data to have $4.14$\% non-Arabic. These turn out to be almost always natural code-switching involving many foreign languages (e.g., English, French, Korean, etc.).

\begin{table}[t]
\centering
\small
\resizebox{0.8\columnwidth}{!}{%
\begin{tabular}{lcHr}
\hline
\textbf{Source}                          & \textbf{Size}  & \textbf{Sentences} & \textbf{Tokens} \\ \hline
AraNews & $8.6$GB                   & $4.2$M           & $847.8$M                                \\
Books      & $650$MB                   & $2.3$M            & $72.5$M                                 \\
El-Khair & $16$GB      & $42$M           & $1.6$B                              \\

Gigawords    & $10$GB                    & $67.4$M           & $1.1$B                              \\

OSIAN     & $2.8$GB                   & $12$M           & $292.6$M                                \\

OSCAR-MSA        & $31$GB                    & $67.3$M           & $3.4$B          \\
OSCAR-Egyptian      &$32$MB & $101.9$M              & $3.8$M                                  \\
Wiki     & $1.4$GB                   & $12.5$M           & $156.5$M                                \\ \hline

\textbf{MSA-Total}                           & \textbf{$\bf 70$}GB & \textbf{$ \bf 205.4$M}                      & \textbf{$\bf 7.1$}B   

\\ \hline

\textbf{Twitter (1.5B)} & $\bf 178$GB                   & $xx$M           & $\bf 21.9$B                               \\\hline
\textbf{ALL} & $\bf 248$GB                   & $xx$M           & $\bf 29.0$B
      \\ 

\hline

\end{tabular}%
}
\caption{The MSA and Twitter  resources used to pre-train AraT5\textsubscript{MSA}, AraT5\textsubscript{TW}, and AraT5.}
\label{tab:art5_data}
\end{table}

\subsection{Pre-Processing and Vocabulary}\label{subsec:Pre-Pr_data}
 We remove diacritics and replace URLs and user mentions with \texttt{<URL>} and \texttt{<USER>}. We also clean the data by removing HTML tags,  elongation, and the hash signs. Further, we  reduce repetitive characters,  emojis, and emoticons to one.  
To create our language model vocabulary, we use SentencePiece \cite{kudo2018subword} to encode text as WordPiece tokens~\cite{smp2016} with $110$K WordPieces. To allow for further pre-training (and/or fine-tuning) on additional languages, we extract our vocabulary as follows: $70$M MSA sentences,  $200$M  Arabic twitter data, $15$M sentences from Wikipedia English, and $5$M sentences from the Wikipedia of  $10$ other languages (Bulgarian, French, German, Greek, Italian, Portuguese, Russian, Spanish, Turkish, Czech).\footnote{The MSA and twitter data are extracted from our training data presented in Section~\ref{subsec:tr_data}.} In \cref{subsec:ARGEN_MT_X}, we describe parallel data from four of these languages on which we fine-tune our models for X$\rightarrow$Arabic MT. Our respective results (reported in Table~\ref{tab:x_to_ar_res_in_voacab}) demonstrate the utility of including foreign vocabulary in our models.

\subsection{AraT5}\label{subsec:AraT5_MSA_TW}
\textbf{Model Architecture.} We leverage our unlabeled MSA  and Twitter data described in \cref{subsec:tr_data} to pre-train three models: \textbf{\texttt{AraT5\textsubscript{MSA}}} on MSA data, \textbf{\texttt{AraT5}\textsubscript{TW}} on twitter data, and \textbf{\texttt{AraT5}} on  both MSA and twitter data using the T5\textsubscript{\textsubscript{Base}} encoder-decoder architecture~\cite{raffel2019exploring}. Each of the encoder and decoder components is similar in size and configuration to  BERT\textsubscript{Base} \cite{devlin2019bert}, with  $12$ layers each with $12$ attention heads, and $768$ hidden units.  In total, this results in a model with $\sim220$ million parameters.\footnote{The output dimensionality is d$_{ff}$ = $3,072$ and  inner dimensionality of d$_{kv}$ = $64$.} 
\textbf{Objective.} \newcite{raffel2019exploring} pre-train T5\textsubscript{\textsubscript{Base}} using a self-supervised (denoising) objective. The main idea is to feed the model with masked (corrupted) versions of the original sentence, and train it to reconstruct the original sequence.  Inspired by BERT's objective~\cite{devlin2019bert}, the  denoising objective ~\cite{raffel2019exploring} works by randomly sampling and dropping out $15$\% of tokens in the input sequence. All consecutive spans of dropped-out tokens are then replaced by a single sentinel token. \textbf{Pre-Training.} For all three of our pre-trained models, we use a learning rate of $0.01$, a batch size of $128$ sequences, and a maximum sequence length of $512$, except for  {AraT5}\textsubscript{TW} where the maximum sequence is $128$.\footnote{We choose the same maximum sequence used in MARBERT~\cite{abdul2020arbert}, the most powerful model trained on Arabic twitter to date~\cite{farha2021benchmarking}.}   We pre-train each model for $1$M steps. Pre-training of each model took\textbf{ $\sim\bf80$ days} on one Google Cloud TPU with $8$ cores (v$3.8$) from TensorFlow Research Cloud (TFRC).\footnote{\href{https://www.tensorflow.org/tfrc}{https://www.tensorflow.org/tfrc}.}  We now introduce  our language generation and understating benchmarks.

\label{sec:our_models}

\section{ARGEN}\label{sec:AraLG}
In order to evaluate our pre-trained language models, we introduce our new benchmark  for Arabic language generation evaluation \textbf{ARGEN}. It includes \textit{19 different datasets} with \textit{59 test splits} and covers \textit{seven} tasks: machine translation (MT), code-switched translation (CST), text summarization (TS), news title generation (NGT), question generation (QG), transliteration (TR), and  paraphrasing (PPH).  As such, ARGEN has wide-coverage both in terms of the number of tasks and datasets. It is also linguistically diverse as it covers both MSA and various Arabic dialects, in addition to \textit{Arabizi} (romanized Arabic in the TS task) and code-switching (in the CST task). We now describe each component of ARGEN.

\subsection{Machine Translation}\label{subsec:ARGEN_MT}
To design the MT component  of ARGEN, \textbf{ARGEN\textsubscript{MT}}, we consolidate $7$ unique datasets with $46$ different test splits. The datasets come from both MSA and Arabic dialects, and range between $600$-$138$K sentences (details in Table~\ref{tab:mt_data_app} in Appendix). We introduce each dataset briefly here.

\subsubsection{Arabic $\rightarrow$ English}

\noindent\textbf{(1) United Nations Parallel Corpus.} \newcite{ziemski2016united} introduce this parallel corpus of manually translated UN documents covering the six official UN languages (i.e., Arabic, Chinese, English, French, Russian, and Spanish). The corpus consists of development and test sets only, each of which comprise $4,000$ sentences that are one-to-one alignments across all official languages.

 \noindent\textbf{(2) IWSLT Corpus.} Several  Arabic-to-English parallel datasets were released  during IWSLT evaluation campaigns~\cite{federico2012overview,cettolo2013report,cettolo2014report,cettolo2016iwslt}. The datasets are mainly extracted from transcriptions of TED talks between 2010 and 2016,  and the QCRI Educational Domain Corpus (QED 2016)~\cite{abdelali2014amara}.




\noindent\textbf{AraBench Datasets.}~\newcite{sajjad2020arabench} introduce AraBench, an evaluation suite for MSA and dialectal Arabic to English MT consisting of \textit{five} publicly available datasets: \textbf{(3) ADPT:} Arabic-Dialect/English Parallel Text ~\cite{zbib2012machine}, \textbf{(4) MADAR:} Multi-Arabic Dialect Applications and Resources dataset~\cite{bouamor2018madar}, \textbf{(5) QAraC:} Qatari-English speech corpus~\cite{elmahdy2014development}, and \textbf{(6) Bible:} The English Bible translated into MSA, Moroccan, and Tunisian Arabic dialects.\footnote{The United Bible Societies \href{https://www.bible.com}{https://www.bible.com}.} For all these datasets, we use the same splits as \newcite{sajjad2020arabench} in our experiments. 

\subsubsection{X $\rightarrow$ Arabic}  \label{subsec:ARGEN_MT_X}

To investigate ability of our models to generate Arabic starting from foreign languages in our vocabulary, we create an X$\rightarrow$Arabic benchmark of four languages (English, French, German, and Russian) by extracting parallel data from OPUS~\cite{OPUS}. For each language, we pick $1$M sentences for training and $5$K sentences for each of development and test splits. This gives us our seventh \textbf{ARGEN\textsubscript{MT}} dataset, which we call \textbf{(7)~OPUS-X-Ara}.





\subsection{Code-Switched Translation}\label{subsec:AraGEN_CST}

There is rising interest in translating \textit{code-switched} data~\cite{nagoudi-cs-2021}. Our purpose here is to translate Arabic text involving code-switching from a foreign language into \textbf{(i)} that foreign language as well as into \textbf{(ii)} MSA. Hence we create \textbf{ARGEN\textsubscript{CST}}, our code-switched translation benchmark component, using \textit{four} sub-test sets. Two of these are \textit{natural} and two are \textit{synthetic}, as follows:

\noindent\textbf{Natural Code-Switched Data. } We create two human written (natural) code-switched parallel datasets: \textbf{(1) ALG-CST.} This is collected from Algerian Twitter and consists of code-switched Arabic-French posts. We translate these manually into monolingual French. \textbf{(2) JOR-CST.} This is collected from Jordanian Twitter and consists of code-switched Arabic-English posts, which we manually translate into monolingual English. Each of ALG-CST and JOR-CST comprises $300$ tweets (total=$600$). Human translation is performed by one native speaker from each dialect with semi-native English/French fluency. 

\noindent\textbf{Synthetic Code-Switched Data.} We use the multi-lingual sequence-to-sequence  model mBART~\cite{liu2020multilingual} to create synthetic code-switched data following~\newcite{jawahar2021exploring}. We exploit the UN multi-parallel data~\cite{ziemski2016united} using the Arabic-English and Arabic-French test splits ($4,000$ sentences each, described in \cref{subsec:ARGEN_MT}) to generate our two code-switched test sets \textbf{(3) MSA-EN} and \textbf{(4) MSA-FR}. In each case, we use mBART to translate $\sim 30\%$ random Arabic n-grams into  the target language (i.e., English or French).

\subsection{Text Summarization}
To build our \textit{text summarization}  benchmark component, \textbf{ARGEN\textsubscript{TS}}, we  use the following:

\noindent\textbf{Essex Arabic Summaries Corpus (EASC).} EASC~\cite{el2010using} contains $153$ Arabic Wikipedia and newspaper articles, each with $5$ human-generated extractive summaries (total=$765$ summaries). The summaries are crowdsourced via Mechanical Turk.\footnote{\href{http://www.mturk.com/}{http://www.mturk.com/}}

\noindent\textbf{WikiLingua.} An abstractive summarization dataset in $18$ languages, including Arabic ~\cite{ladhak-wiki-2020}. 
It contains articles and their summaries from WikiHow.\footnote{\href{http://www.wikihow.com}{http://www.wikihow.com}} 
 The Arabic part includes summaries for $29.2$K articles, which we split into 80\% Train ($23.4$K), 10\% Dev ($2.9$K), and 10\% Test ($2.9$K).


\subsection{News Title Generation}
The purpose of the \textit{news title generation (NTG)} task is to produce proper news article titles~\cite{liang2020xglue}. We introduce NTG as a \textit{new} task for Arabic language generation. Given an article, a title generation model needs to output a short grammatical sequence of words suited to the article content. For this, we introduce \textbf{{ARGEN\textsubscript{NTG}}}, a novel NTG dataset exploiting $120$K articles along with their titles extracted from AraNews~\cite{nagoudi2020machine}.\footnote{We ensure no overlap exists between ARGEN\textsubscript{TG} and the AraNews data we use to pre-train our language models (described in~\cref{sec:our_models}).} We only include titles with at least three words in this dataset. We  split ARGEN\textsubscript{NTG} data  into 80\% Train ($93.3$K), 10\% Dev ($11.7$K), and 10\% Test ($11.7$K). Details about ARGEN\textsubscript{NTG} are in Table~\ref{tab:arnews_tg} (Appendix). A sample of a news article from our Test split and example titles generated by our models are in Table~\ref{tab:ngt_examples} (Appendix).

\subsection{Question Generation}
In the \textit{question generation (QG)} task, a question is produced for a passage~\cite{gehrmann2021gem}.  Given  the absence of an Arabic QG  dataset, we create a new Arabic QG dataset (\textbf{ARGEN\textsubscript{QG}}) using a publicly available Arabic question answering (QA) resource. We follow \newcite{kriangchaivech2019question} who train a model to generate simple questions relevant to passages and answers extracted from  SQuAD \cite{rajpurkar2016squad}. In our case, we build ARGEN\textsubscript{QG} by extracting $96$K (passage, answer, and  question) triplets from \textbf{(1)} The Arabic QA dataset  ARCD~\cite{mozannar2019neural}, and \textbf{(2)}  three multi-lingual  QA datasets: XTREME benchmark~\cite{pmlr-v119-hu20b},  MLQA~\cite{lewis2019mlqa}, XQuAD~\cite{artetxe2020cross}, and  TyDi QA~\cite{artetxe2020cross}.

\subsection{Paraphrasing}
\label{subsec:paraphrasing}
The main goal of this task is to produce for a given Arabic sentence a \textit{paraphrase} with the same meaning. In order to build our paraphrasing  benchmark component (\textbf{ARGEN\textsubscript{PPH}}), we  use the following three datasets:

\noindent\textbf{AraPara.} We introduce AraPara, a new multi-domain Arabic paraphrasing dataset we create using English-Arabic parallel OPUS data~\cite{OPUS}. AraPara covers several domains such as news, religion, politics, movies,  and technology. To create a high quality machine generated paraphrase dataset, we follow four careful steps involving human validation (more details are offered in Appendix~\ref{sec:ARGEN_parah_app}). AraPara consists of $122$K paraphrase pairs. We only use AraPara for model development, and hence we split it into $116$K Train and $6$K Dev.


    




\noindent\textbf{Arabic SemEval Paraphrasing (ASEP).}
We also create a new Arabic paraphrasing dataset using three existing Arabic semantic similarity datasets released during SemEval 2017~\cite{cer2017semeval}. These are MSR-Paraphrase ($510$ pairs), MSR-Video ($368$ pairs), and SMTeuroparl ($203$ pairs). The pairs are labeled with a similarity score on a scale from $0$ to $5$. For our purpose, we only keep sentence pairs with a semantic similarity score $\geq 3.5$ which gives us $603$ pairs. We merge and shuffle all three ASEP datasets for our use. 

\noindent\textbf{Arabic Paraphrasing Benchmark (APB).} APB is created by~\newcite{alian2019towards}. It consists of $1,010$ Arabic sentence pairs that are collected from different Arabic books. Paraphrasing was performed manually using six transformation procedures (i.e., addition, deletion, expansion, permutation, reduction, and replacement).

\subsection{Transliteration.}

\textit{Transliteration} involves mapping a text written with orthographic symbols in a given script into another~\cite{beesley1998romanization}. We use the \textit{BOLT Egyptian Arabic SMS/Chat and Transliteration dataset}~\cite{song2014collecting},\footnote{\href{https://catalog.ldc.upenn.edu/LDC2017T07}{https://catalog.ldc.upenn.edu/LDC2017T07}} a collection of  naturally-occurring chat and short messages (SMS) from Egyptian native speakers.  The messages (sources) were natively written in either romanized Arabizi or Egyptian Arabic orthography. The target is the Egyptian transliteration of these message.\footnote{Some transliteration sequences involve code mixing between Egyptian Arabic and
English.}  For experiments, we use the same split proposed by~\newcite{shazal2020unified} ($58.9$K for Train and $5.4$K  for Dev and Test each). We refer to this dataset as \textbf{ARGEN\textsubscript{TR}}.

\label{sec:AraGen}

\section{Evaluation on ARGEN}\label{sec:EVAL}

\noindent\textbf{Baselines and Procedure.}
For all tasks, we compare our models to models fine-tuned with mT5 using the same training data. In addition, for MT, we compare to a vanilla sequence-to-sequence (S2S) Transformer~\cite{vaswani2017attention} trained from scratch as implemented in Fairseq~\cite{ott2019fairseq}. For all models and baselines, across all tasks, we identify the best model on the respective Dev data and blind-test it on Test data. As a rule, we report on both Dev and Test sets. All our Dev results are in Section~\ref{sec:Arabic_LG_eval_app} in the Appendix.

\begin{table*}[]
\centering
 \renewcommand{\arraystretch}{1.2}
\resizebox{.9\textwidth}{!}{%
\begin{tabular}{lllHrrr|rrr|r}
\toprule
\multicolumn{2}{c}{ \textbf{ Dataset}   }    & \textbf{ Test Split}  & \textbf{ SOTA}    & \textbf{\colorbox{red!15}{S2S\textsubscript{2M}}}  & \colorbox{red!15}{\textbf{S2S\textsubscript{10M}}}&  \colorbox{red!10}{\textbf{  mT5}}  & \colorbox{green!10}{\textbf{ AraT5\textsubscript{Tw}}} & \colorbox{green!13}{\textbf{  AraT5\textsubscript{MSA}}} & \colorbox{green!13}{\textbf{  AraT5} } & \textbf{ \colorbox{gray!7}{SOTA}}\\
\toprule

&\multirow{2}{*}{{\textbf{ADPT}\textsuperscript{$\dagger$}}}   &  Lev &  $\bf10.80$&$4.30$&$6.20$&$8.33$&$8.32$&$\bf8.52$&$8.42$&  $10.80$     \\

&    &  Egy &  $\bf14.00$&$5.21$&$8.9$&$12.57$&$11.25$&$12.38$&$\bf12.92$ &  $14.00$   \\
\cdashline{2-11}
&\multirow{2}{*}{{\textbf{Bible I}} }  & \multirow{1}{*}{Tun.}     & $7.00$&$4.12$&$4.44$&$8.08$&$5.86$&$\bf8.52$&$7.94$& $7.00$         \\
 & &     \multirow{1}{*}{Mor.}      &$4.20$&$2.60$&$2.80$&$7.21$&$4.69$&$\bf7.83$&$6.82$&$4.20$          \\ \cdashline{2-11}

& \multirow{5}{*}{\textbf{MADAR I}\textsuperscript{$\dagger$}} & \multirow{1}{*}{Egy.} &$\bf28.90$&$17.25$&$17.71$&$24.44$&$21.75$&$\bf24.98$&$24.66$ &$28.90$   \\
& & \multirow{1}{*}{{Qat.}}                                   &$\bf27.60$&$15.98$&$17.92$&$23.72$&$22.23$&$\bf24.00$&$23.92$&$27.60$   \\  
& &\multirow{1}{*}{{Leb.}}                                      & $\bf17.00$&$12.15$&$10.14$&$14.61$&$12.25$&$\bf14.92$&$14.18$  & $17.00$\\ 
& &\multirow{1}{*}{{Tun.}}                                        & $\bf11.40$&$8.49$&$8.57$&$10.12$&$9.09$&$\bf10.18$&$9.60$  & $11.40$    \\  
& &\multirow{1}{*}{{Mor.}}                                        & $\bf14.70$&$11.07$&$11.83$&$16.61$&$12.37$&$\bf16.99$&$16.82$ & $14.70$  \\   \cdashline{2-11}

\multirow{18}{*}{{\textbf{DIA}}}  &\multirow{20}{*}{{\textbf{MADAR II}\textsuperscript{$\dagger$}}} 
& Egy-Alex.   &  $28.90$ / $29.34$ &$19.01$&$19.74$&$29.34$&$24.79$&$\bf29.87$&$29.02$  &  $28.90$   \\
& &  Egy-Asw.  & $\bf26.30$&$16.37$&$16.95$&$23.01$&$19.52$&$\bf23.41$&$22.06$ & $26.30$\\
& &  Sud-Kha.  &$\bf36.70$&$24.97$&$25.65$&$30.87$&$28.13$&$\bf31.39$&$30.65$       &$36.70$       \\ 
& &  Yem-San.  & $\bf29.90$&$19.62$&$20.35$&$24.87$&$23.19$&$\bf26.10$&$25.73$    & $29.90$     \\
& &  Oma-Mus. & $\bf39.50$&$29.12$&$30.66$&$33.74$&$32.15$&$\bf34.62$&$34.18$  & $39.50$      \\
& &  KSA-Riy. &  $\bf40.70$&$26.14$&$26.66$&$33.54$&$30.81$&$\bf33.86$&$33.59$   &  $40.70$    \\
& &  KSA-Jed. & $\bf27.40$&$16.08$&$17.21$&$\bf23.57$&$20.91$&$23.45$&$23.11$  & $27.40$      \\
& &  Iraq-Bag. & $\bf28.30$&$15.98$&$19.09$&$22.92$&$20.84$&$23.24$&$\bf22.52$  & $28.30$     \\ 
& &  Iraq-Bas. & $\bf27.70$&$16.46$&$17.12$&$\bf22.94$&$20.47$&$22.61$&$22.00$  & $27.70$    \\ 
& &  Iraq-Mos.  & $\bf30.00$&$18.25$&$19.14$&$23.69$&$21.95$&$\bf24.41$&$23.12$  & $30.00$      \\ 

& & Pal-Jer. &  $\bf27.0$&$15.18$&$16.06$&$24.61$&$20.91$&$\bf24.95$&$24.45$    &  $27.00$ \\
& & Jor-Amm. &  $\bf30.00$&$18.68$&$18.86$&$26.45$&$22.92$&$\bf26.78$&$25.26$     &  $30.00$   \\
& & Jor-Salt. & $\bf29.60$&$17.14$&$17.78$&$26.04$&$23.05$&$\bf26.56$&$26.05$   & $29.60$  \\
& &  Syr-Dam. &  $\bf25.90$&$13.63$&$14.83$&$21.93$&$18.55$&$\bf22.54$&$21.80$    &  $25.90$   \\
& &  Syr-Alep. &  $\bf26.40$&$14.16$&$15.27$&$22.39$&$19.55$&$22.91$&$\bf23.26$    &  $26.40$  \\  
& &   Alg-Alg.  & $\bf17.30$&$13.94$&$14.24$&$16.97$&$14.26$&$\bf17.46$&$16.62$    & $17.30$\\ 
& &  Lyb-Trip. & $\bf22.80$&$14.49$&$15.44$&$20.17$&$17.56$&$\bf20.31$&$19.85$  & $22.80$  \\ 
& &  Lyb-Beng. & $\bf28.40$&$19.02$&$19.32$&$25.50$&$23.39$&$25.46$&$\bf25.54$  & $28.40$ \\ 
& &   Tun-Saf & $\bf10.80$&$7.89$&$8.57$&$9.26$&$8.15$&$\bf9.94$&$9.60$     & $10.80$ \\ 
& &  Mor-Fes & $20.90$&$15.09$&$15.59$&$22.81$&$17.33$&$\bf23.33$&$21.97$   & $20.90$ \\ \cdashline{2-11}
&\textbf{QAraC}\textsuperscript{$\dagger$}& Qatar & $\bf11.9$&$10.33$&$10.47$&$\bf11.84$&$11.11$&$11.42$&$10.57$   & $11.90$ \\ 
\cline{2-11}
 &  \colorbox{blue!9}{\textbf{\textit{Average DIA}}} &    \textbf{\textit{ }} & $\bf11.9$&$14.75$&$15.58$&$20.66$&$18.28$&\colorbox{blue!9}{$\bf21.02$}&$20.49$   & $23.49$ \\ 

\toprule
 & \multirow{2}{*}{{\textbf{Bible II}\textsuperscript{$\dagger$}}}  &  Test 1   &$\bf17.00$&$10.44$&$10.86$&$15.58$&$13.04$&$\bf16.38$&$15.71$  &$17.00$   \\
  &   & Test 2   & $\bf12.80$&$5.55$&$6.20$&$12.14$&$9.27$&$\bf12.53$&$11.64$  & $12.80$          \\ \cdashline{2-11}
\multirow{6}{*}{{\textbf{MSA}}}   & \textbf{MADAR I}\textsuperscript{$\dagger$}  & MSA & $\bf11.90$&$10.33$&$10.47$&$\bf11.84$&$11.11$&$11.42$&$10.57$ & $11.90$    \\   \cdashline{2-11}
  & \multirow{8}{*}{\textbf{IWSLT}\textsuperscript{$\ddagger$}}  & 
   {TED10} &  $-$& $24.12$&$25.13$&  $28.02$ &	$27.35$ &	$\bf28.64$ &	$28.32$ &  $28.00$         \\
    & & {TED11} & $-$&$23.96$&$25.01$&  $28.89$ &	$28.03$ &	$\bf29.93$ &	$27.34$ &   $32.80$               \\
  & & {TED12} & $-$&$28.34$&$28.98$&  $33.77$ &	$32.74$ &	$\bf35.07$ &	$34.238$ & $36.50$         \\ 
  &  & {TED13}&  $-$&$24.19$&$25.02$& $27.12$ &	$27.52$ &	$\bf27.95$ &	$27.52$ &  $37.40$                \\ 
  
    &  & {TED14}& $-$&$25.64$&$26.48$& $29.85$ &	$28.64$ &	$\bf30.94$ &	$30.06$ &   $31.70$               \\

 &  &  {TED15} &  $\bf34.1$&$27.68$&$28.73$&$29.39$&$28.2$&$30.37$&$\bf30.45$  &  $34.10$                \\
  & & {TED16}  &   $\bf31.80$&$25.71$&$25.77$&$28.39$&$27.03$&$\bf29.37$&$29.18$    &   $31.80$             \\ 
  &  & {QED16}  &  $\bf28.10$&$19.44$&$19.90$&$\bf21.09$&$18.55$&$20.98$&$19.11$  &  $28.10$              \\ \cdashline{2-11}
    & \textbf{UN}\textsuperscript{$\dagger\dagger$}  &AR-EN&$\bf56.90$&$52.54$&$53.12$&$52.38$&$51.48$&$53.29$&$\bf52.96$  &$56.90$              \\

\cline{2-11}
 &  \colorbox{blue!10}{\textbf{\textit{Average MSA}}} &    \textbf{\textit{ }} & $\bf25.75$&$23.54$&$24.19$&$27.03$&$25.43$&\colorbox{blue!9}{$\bf27.77$}&$26.98$ &$30.63$\\

\toprule
 &  \colorbox{blue!9}{\textbf{\textit{Average All}}} &    \textbf{\textit{}} & $\bf25.75$&$19.14$&$19.89$&$23.84$&$21.85$&\colorbox{blue!9}{$\bf24.39$}&$23.74$ &$27.06$\\
\toprule

\end{tabular}
}

\caption{English  to  Arabic results in BLEU using ARGEN\textsubscript{MT}  datasets. \textbf{\colorbox{red!20}{Baseline I}:} Sequence-to-Sequence Transformer models trained from scratch on $2$M and $10$M parallel sentences. \textbf{\colorbox{red!7}{Baseline II}:} mT5~\cite{xue2020mt5}. \textbf{\colorbox{green!10}{Our models}:} ArT5\textsubscript{Tweet},  ArT5\textsubscript{MSA}, ArT5. \textbf{\colorbox{gray!7}{SOTA}:} \textsuperscript{$\dagger$}~\newcite{sajjad2020arabench} trained on $\sim42$M sentences,  \textsuperscript{$\ddagger$}~\newcite{durrani2017qcri} trained on $\sim59$M sentences, \textsuperscript{$\dagger\dagger$}~\newcite{junczys2016neural} trained on $\sim12$M sentences. }

\end{table*}\label{tab:RES_MT36}
\subsection{Machine Translation.}
We train two S2S Transformers models on $2$M (S2S\textsubscript{2M}) and $10$M (S2S\textsubscript{10M}) MSA-English parallel sentences extracted from OPUS. We take these two models as our \textbf{baseline~I}. We also fine-tune our three models as well as mT5 on the same OPUS $2$M MSA-English parallel sentences used for baseline~I. Fine-tuned mT5 is our second baseline \textbf{baseline II}.%

\noindent \textbf{Arabic $\rightarrow$ English.} Results of ARGEN\textsubscript{MT} are reported in Table~2. Results show that our models achieve best BLEU score in  $37$ out of the $42$ tests splits. AraT5\textsubscript{MSA} acquires best results in $32$ of these test splits, outperforming all the baselines (S2S\textsubscript{2M}), (S2S\textsubscript{10M}), and mT5 with  +$5.25$, +$4.99$, and +$0.45$ BLEU points. These results are striking since our language models are pre-trained on Arabic data only (although they include English vocabulary and marginal amounts of code-switching; see~\cref{subsec:tr_data}). In other words, even under this arguably \textit{zero-shot} setting,\footnote{At best, this can be viewed as \textit{few-shot} pre-training.} the models perform very well. In addition, our AraT5 model outperforms even the S2S model trained with $5$X more data. For completeness, we also provide the current SOTA on each of our datasets. We do not compare our results to SOTA since these are acquired by models fine-tuned on much larger datasets than ours. For example, ~\newcite{sajjad2020arabench} exploit $\sim 42$M parralel sentences to train their models. To limit GPU needs during our experiments, especially given the time-consuming fine-tuning process typical of T5 models, we do not fine-tune the models on the full amounts of available parallel data. However, in the future we plan to compare our models under the full data setting.

\noindent \textbf{X $\rightarrow$  Arabic.}
Our language models are not pre-trained on foreign data, but we include vocabulary from $11$ foreign languages. Our X $\rightarrow$ Arabic experiments here are hence zero-shot (from the perspective of pre-training). Table~\ref{tab:x_to_ar_res_in_voacab} shows the results of AraT5\textsubscript{MSA} and mT5 on OPUS-X-Ara.\footnote{To limit GPU time, we fine-tune only AraT5\textsubscript{MSA} model on the X$\rightarrow$Arabic direction since it performed best on Arabic$\rightarrow$English section above.}
 We observe that our model outperforms mT5 in the four X $\rightarrow$ Arabic sub-tasks  with an average of +$1.12$ and +$0.86$ BLEU points on Dev and Test, respectively.

\subsection{Code-Switched Translation.}
For this task, we test on the two natural code-switched translation (CST) test sets that we manually created, ALG-FR$\rightarrow$FR and JOR-EN$\rightarrow$EN. We also evaluate on our two synthetic CST datasets, MSA-EN and MSA-FR, one time with EN/FR as target (e.g., MSA-EN$\rightarrow$EN) and another with MSA as target (e.g., MSA-EN$\rightarrow$MSA). We fine-tune our three pre-trained models as well as mT5 on the OPUS-X-Ara segments involving English and French (each with $1$M parallel sentences, described in~\cref{subsec:ARGEN_MT_X}), in both directions. Since these MT models are only fine-tuned on parallel monolingual data, we refer to these experiments as \textit{zero-shot}. We test these models on both our natural and synthetic code-switched data (described in~\cref{subsec:AraGEN_CST}). We report results in Table~\ref{tab:cs_TEST}. Our models achieve best results in one out of the two natural test sets (with +$4.36$ BLEU points on ALG-FR) and \textit{all four} synthetic test sets (e.g., +$4.55$ BLEU points on MSA-EN$\rightarrow$MSA). \textit{These results clearly show our models' remarkable language generation ability especially in the Arabic direction.}


\begin{table}[ht]
\centering
 \renewcommand{\arraystretch}{1.5}
\resizebox{1\columnwidth}{!}{%
\begin{tabular}{llcccc}
\toprule 
\textbf{Dataset}     & \textbf{~~~~~~~~~~~Split} & \textbf{mT5} & \textbf{AraT5\textsubscript{Tw}} & \textbf{AraT5\textsubscript{MSA}} & \textbf{AraT5\textsubscript{}} \\\toprule
\multicolumn{1}{c}{\multirow{2}{*}{Natural}}
   & ALG-FR $\rightarrow$  FR                                & $23.83$	& $\bf28.19$	& $26.27$	& $26.17$                                     \\ 
   & JOR-EN $\rightarrow$  EN                                  & $\bf23.06$	& $21.60$	& $21.58$	& $20.45$                                     \\ 
   \hline\hline
\multicolumn{1}{c}{\multirow{4}{*}{Synthetic}}
   & MSA-FR $\rightarrow$  FR                                   & $12.76$	& $10.57$	& $\bf13.78$	& $13.25$                                     \\  
   &  MSA-EN $\rightarrow$  EN                                & $11.06$	& $8.99$	& $\bf11.53$	& $11.42$                                     \\    \cdashline{2-6}
	
      & MSA-FR $\rightarrow$  MSA                                   & $12.93$	& $12.14$	& $\bf14.39$	& $13.92$                                     \\  
   &  MSA-EN $\rightarrow$  MSA                                & $19.82$	& $18.43$	& $23.89$	& $\bf24.37$                                     \\       
  \toprule
\end{tabular}%
}
\caption{Performance of our models on \textbf{ARGEN\textsubscript{CS}}.} \label{tab:cs_TEST}
\end{table}

\vspace*{-\baselineskip}



\begin{table}[t]
\centering
\resizebox{0.9\columnwidth}{!}{%
\begin{tabular}{lHcccc}
\toprule 
\multirow{2}{*}{\textbf{Dataset}} & \multirow{2}{*}{\textbf{S/M}} & \multicolumn{2}{c}{\textbf{DEV}}              & \multicolumn{2}{c}{\textbf{TEST}} \\ \cline{3-6}
                      &                                    & \multicolumn{1}{c}{\textbf{mT5}} & \textbf{AraT5\textsubscript{MSA}} & \textbf{mT5}    & \textbf{AraT5\textsubscript{MSA}}     \\ 
\toprule 

EN $\rightarrow$ AR& S                              & $13.60$&$\bf15.72$&$17.80$&$\bf18.58$\\

DE $\rightarrow$ AR   & S                              & $12.88$&$\bf13.74$&$11.92$&$\bf12.80$\\
FR $\rightarrow$ AR                                      & S       &  $17.52$&$\bf17.96$&$18.61$&$\bf18.99$\\ 

RU $\rightarrow$ AR                                    & S                                & $26.78$&$\bf27.87$&$26.63$&$\bf28.01$\\       

\hline
\multirow{1}{*}{\textbf{Average}}                                    & S                               &$17.70$	&$\bf18.82$&	$18.74$&	$\bf19.60$\\
\toprule
\end{tabular} }
\caption{Performance of MT models on OPUS-X-Ara.}

\end{table}\label{tab:x_to_ar_res_in_voacab}


\subsection{Text Summarization}
For the two ARGEN\textsubscript{ST} datasets, we fine-tune and identify the best model on the Train and Dev splits of WikiLingua~\cite{ladhak-wiki-2020} and test on all EASC and the Test of WikiLingua. We report different ROUGE scores~\cite{lin2004rouge} in Table~\ref{tab:Arbechgen_summary_TEST}. As the Table shows, AraT5\textsubscript{Tw} acquires best results on WikiLingua data, while mT5 outperforms us on EASC (we hypothesize since EASC is older data that is likely part of the mC4 on which mT5 was pre-trained). \textit{On both datasets, we establish new SOTA} (both with our pre-trained models and mT5).

\begin{table}[ht]
\centering
\resizebox{1\columnwidth}{!}{%
\begin{tabular}{lccccc}
\toprule 
\textbf{Dataset}     & \textbf{Metric} & \textbf{mT5} & \textbf{AraT5\textsubscript{Tw}} & \textbf{AraT5\textsubscript{MSA}} & \textbf{AraT5\textsubscript{}} \\\toprule

\multicolumn{1}{c}{\multirow{3}{*}{EASC}}
   & Rouge1                                  & $\bf62.98$	& $60.74$	& $59.54$	& $54.61$                                     \\ 
   & Rouge2                                 &  $\bf51.93$	& $48.89$	& $47.37$	& $43.58$                                    \\
    & RougeL                                  & $\bf62.98$	& $60.73$	& $59.55$	& $54.55$                                     \\ 
   \hline
\multicolumn{1}{c}{\multirow{3}{*}{WikiLin.}}
   & Rouge1                                  & $71.63$	&	$\bf74.61$& 	$72.64$& 	$73.48$                                     \\ 
   & Rouge2                                 &  $63.60$                      &$\bf67.00$& 	$64.21$	& $65.09$                                    \\
    & RougeL                                  & $71.56$	&	$\bf74.52$	& $72.57$	& $73.37$                                    \\ 
  \toprule
\end{tabular}%
}
\caption{Performance of summarization models on Test. We consider mT5 as SOTA for WikiLin,  and  ~\newcite{alami2021unsupervised} (ROUGE1=$59.17$) for EASC.} \label{tab:Arbechgen_summary_TEST}
\end{table}
\subsection{News Title and Question Generation}
 For both tasks, we fine-tune all our models on the Train splits of ARGEN\textsubscript{NTG} and ARGEN\textsubscript{QG}, respectively. As Table~\ref{tab:NGT_QA_TEST} shows, \textit{all} our models outperform mT5 on each of the two tasks. AraT5\textsubscript{MSA} excels with $20.61$\% BLEU on ARGEN\textsubscript{NTG} and AraT5 is at $16.99$\% on ARGEN\textsubscript{QG}.

\subsection{Paraphrasing and Transliteration}
For the \textit{paraphrasing} task, we fine-tune and validate on our new AraPra dataset and blind-test on both APB and ASEP datasets (described in\cref{subsec:paraphrasing}). As Table~\ref{tab:NGT_QA_TEST} shows, AraT5\textsubscript{MSA} is best on APB ($17.52$ BLEU) and ASEP ($19.38$ BLEU).
For \textit{transliteration}, we fine-tune our models on the Train split of ARGEN\textsubscript{TR}. As Table~\ref{tab:NGT_QA_TEST} shows, each of AraT5\textsubscript{MSA} and AraT5 outperform mT5. Notably, AraT5\textsubscript{MSA} is at $65.88$ BLEU, outperforming previous SOTA~\cite{shazal2020unified} by $7.1$ points.

\begin{table}[ht]
\centering
 \renewcommand{\arraystretch}{1.2}
\resizebox{1\columnwidth}{!}{%
\begin{tabular}{lcccc}
\toprule 
\textbf{Dataset}      & \textbf{mT5} & \textbf{AraT5\textsubscript{Tw}} & \textbf{AraT5\textsubscript{MSA}} & \textbf{AraT5} \\\toprule
ARGEN\textsubscript{NTG}            & $19.49$&$20.00$	&$\bf20.61$	&$20.51$           \\\cline{2-5}

ARGEN\textsubscript{QG}

                               & $15.29$ &	$12.06$	&$14.18$&	$\bf16.99$         \\\cline{2-5}

ARGEN\textsubscript{TR}
                            & $60.81$ & $59.55$ &      $\bf65.88$ &  $62.51$         \\\cline{2-5}
    
ARGEN\textsubscript{PPH} I                             & $19.32$ &   $18.17$ &     $\bf19.38$ &  $19.03$         \\
    
ARGEN\textsubscript{PPH} II                                 & $19.25$ &      $17.34$ &   $\bf19.43$ &  $18.42$         \\
\toprule

\end{tabular}%
}
\caption{Performance of our models on title, question generation, transliteration, and paraphrasing tasks in BLEU. \textbf{ARGEN\textsubscript{PPH} I and II}: results on ASEP and APB paraphrase datasets, respectively.  We  consider  mT5  as  SOTA for NTG, QG, and PPH \textbf{ARGEN\textsubscript{NTG}}, \textbf{ARGEN\textsubscript{QG}}, and \textbf{ARGEN\textsubscript{PPH}}. For \textbf{ARGEN\textsubscript{TR}}, SOTA is~\newcite{shazal2020unified} (BLEU=65.88). }
 \label{tab:NGT_QA_TEST}
\end{table}

\subsection{Evaluation on Arabic NLU}
We also evaluate our new pre-trained models on the recently proposed Arabic language understanding and evaluation benchmark, ARLUE~\cite{abdul2020arbert} that involves six cluster tasks (i.e., sentiment analysis, social meaning, topic classification, dialect identification, named entity recognition, and question answering). Our models establish new SOTA on the benchmark with an \textit{ARLUE score of $77.52$} vs. the previous SOTA of $76.53$, reported by ARLUE authors. We provide results of this set of experiments in Appendix~\ref{sec:arbench}.

\label{sec:eval}
\section{Analysis and Discussion}\label{sec:ana_disc}

\subsection{Multilingual vs. Dedicated Models.} Our results confirm the utility of dedicated language models as compared to multilingual models such as mT5 ($101+$ languages). Our AraT5 model outperforms mT5, even though it is pre-trained with $49\%$ less data (see~\cref{subsec:tr_data}). One reason might be that massively multilingual models are more prone to suffering from capacity issues. Data quality is another challenge for multilingual models. As pointed out earlier,~\newcite{caswell2021quality} find systematic issues with data representing several languages (including Arabic) in the mC4 dataset on which mT5 is pre-trained. We perform a data quality study confirming the findings of~\newcite{caswell2021quality}. We also find Arabic mC4 data to be less geographically diverse than our Twitter pre-training data (described in \cref{subsec:tr_data}). Our mC4 data study is in Appendix~\ref{sec:ara-mc4-study}.  

\noindent{\textbf{Code-Switching.}} We also study code-switching in both our Twitter dataset and the Arabic part of mC4. We find that while our Twitter data involves natural code-switching ($\sim4\%$ of sequences), code-switching in Arabic mC4 is very rare. This explains the strong performance of our AraT5\textsubscript{Tw} model on the natural code-switched translation data on French. We conjecture that mT5 good performance on English code-switched data is due to it being pre-trained on very large amounts of English rather than natural code-switching.

\subsection{\textbf{Effect of Sample Length on MT.}}

 We were inquisitive how MT models fine-tuning our pre-trained language models compare to mT5 under different length conditions. For this, we \textbf{(1)} merge all MSA and dialectal Test datasets in our Arabic$\rightarrow$English experiments to form a single dataset that we then \textbf{(2)} split into three bins/Test sets based on sentence length as shown in Table~\ref{tab:seq_length_test}. As the Table shows, our AraT5\textsubscript{MSA} outperform mT5 in \textit{all} but one condition (where our model acquires marginally less performance). We also performed similar evaluation on the merged Dev sets of all MSA and dialectal Arabic MT datasets in the Arabic$\rightarrow$English direction. We do not show related results here, but we note our AraT5\textsubscript{MSA} outperforms mT5 on \textit{all} conditions.


\subsection{\textbf{Qualitative Analysis.}}
We also perform qualitative analyses of the outputs of several of our models, including as to length of MT source data (Appendix~\ref{sec:qualitative_analysis_app}). In particular, our analyses are for the following tasks: machine translation, code-switched translation,  paraphrasing, transliteration,   and news title generation.
\noindent\textbf{MT Model.} Table~\ref{tab:msa_dia_examples_MT} (Appendix) shows three examples of Arabic$\rightarrow$English MT models. Sentence (1) is in \textbf{MSA source}, sentence (2) is in Levantine Arabic source, and sentence (3) is in Egyptian source. In all three examples, one or more of our models generate(s) more fluent translations than mT5. This includes ability of our models to translate dialectal sentences where mT5 seems to struggle (e.g., mT5 is not able to translate the equivalents of ``drive" from Egyptian Arabic).

\noindent\textbf{Code-Switched Translation Model.} Table~\ref{tab:cs_examples_MT} shows two code-switched examples from ARGEN\textsubscript{CS}. Sentence (1) is Algerian dialect at source translated into French, while sentence (2) is Jordanian dialect translated into English. In both cases, our models not only handle the dialects but also their use in code-switched contexts better than mT5.
\begin{table}[t]
\centering
 \renewcommand{\arraystretch}{1.7}
\resizebox{1\columnwidth}{!}{%
\begin{tabular}{ll}
\toprule

\textbf{(1) Source:} &  J'aime une vidéo Episode 1 - \<  نسيبتي العزيزة > 4
\textbf{~~~\colorbox{blue!10}{:ALG-FR}}

\\\cline{2-2}

\textbf{Target:} & \textbf{\colorbox{blue!10}{FR}:}    J' aime une vidéo Episode 1 - ma chère belle-mère  4

\\\cline{2-2}
\textbf{mT5} & 
   
 \colorbox{green!10}{J' aime une} \colorbox{red!10}{v-} \colorbox{green!10}{Chère} \colorbox{red!10}{nièce}   \colorbox{green!10}{4.} 
\\\cdashline{2-2}

\textbf{\textbf{AraT5\textsubscript{Tw}}}     & \colorbox{green!10}{J'aime une vidéo Episode 1 - ma chère} \colorbox{red!10}{tante} \colorbox{green!10}{4.}  
\\\cdashline{2-2}
\textbf{\textbf{AraT5\textsubscript{MSA}}}  & \colorbox{green!10}{ J'aime une vidéo 1 - Ma chère} \colorbox{red!10}{\textbf{sœur} } \colorbox{green!10}{4.} 

  \\\cdashline{2-2}
\textbf{\textbf{AraT5}}  & \colorbox{green!10}{ J'aime une vidéo 1 - Ma chère} \colorbox{red!10}{\textbf{bébé} }                                                              \\
\hline

\textbf{(2) Source:} &  
\< وهاد شيء بائس حقيقة>  comfort zone  \<بطلة العالم في ال> 
 \textbf{~~~\colorbox{blue!10}{:JOR-EN}}                        

\\\cline{2-2}

\textbf{Target:} & \textbf{\colorbox{blue!10}{EN}:}  The world champion in the comfort zone and this is really miserable 

\\\cline{2-2}
\textbf{mT5} & 
 \colorbox{green!10}{the world} \colorbox{red!10}{world} \colorbox{green!10}{champion in comfort zone, and that's really a \textbf{bad} thing.}  
\\\cdashline{2-2}

\textbf{\textbf{AraT5\textsubscript{Tw}}}     & 
\colorbox{green!10}{the world \textbf{hero} in comfort zone  and it's really a \textbf{miserable} thing. }   
\\\cdashline{2-2}
\textbf{\textbf{AraT5\textsubscript{MSA}}}  &                                         \colorbox{green!10}{world champion in comfort zone, and that's really a \textbf{bad} thing. }                         
  \\\cdashline{2-2}
\textbf{\textbf{AraT5}}  &             \colorbox{green!10}{the world's}\colorbox{red!10}{ the world's} \colorbox{green!10}{\textbf{hero} in the comfort zone, and it's a really \textbf{bad} thing.}                                                                         
\\

\toprule 
\end{tabular}%
}
\caption{CS sentences with their English/French translations using our Models and mT5. Data samples are extracted from the Dev datasets. \colorbox{green!10}{\textbf{Green}} refers to good translation. \colorbox{red!10}{\textbf{Red}} refers to problematic translation.}
    \label{tab:cs_examples_MT}
\end{table} 

\noindent\textbf{Paraphrasing, Transliteration, and Title Generation.} Each of Tables~\ref{tab:php_examples},~\ref{tab:tr_examples}, and~\ref{tab:ngt_examples} (Appendix~\ref{sec:qualitative_analysis_app}) shows two output samples from our paraphrasing, transliteration, and title generation models, respectively. In each case, the samples are high-quality, informative, and fluent. Our paraphrase samples also tightly capture the meaning of the source sentences.

\label{sec:discussion}
\section{Related Work}\label{sec:RW}
\textbf{Multilingual LMs.} \noindent\textit{\textbf{mBERT}} is the multilingual version of BERT~\cite{devlin2019bert}, which is an encoder model with bidirectional representations from Transformers trained with a denoising objective. 
 mBERT is trained on Wikipedia for $104$ languages, including Arabic. 
\noindent\textit{\textbf{XLM-R}} ~\cite{conneau2019unsupervised} is also a Transformer-based multilingual masked language model pre-trained on more than $2$TB of CommonCrawl (CC) data in $100$ languages, including Arabic ($2.9$B tokens). 
XLM-R model uses the same masking objective as BERT, but not the next sentence prediction. 
\noindent\textit{\textbf{mT5}}~\cite{xue2020mt5} is the multilingual version of \textbf{T}ext-\textbf{t}o-\textbf{T}ext \textbf{T}ransfer \textbf{T}ransformer model (T5)~\cite{raffel2019exploring}. T5 is an encoder-decoder Transformer similar in configuration and size to a BERT$_{Base}$. 
It is trained on mC4, which is  $\sim26.76$TB for $101$ languages generated from $71$ CC dumps.



\noindent\textbf{Arabic LMs.} \noindent\textit{\textbf{AraBERT}}~\cite{antoun2020arabert} is an Arabic pre-trained language model based on the  BERT\textsubscript{Base} architecture with 24GB of MSA data. 
\noindent\textit{\textbf{ARBERT}} and \noindent\textit{\textbf{MARBERT}}~\cite{abdul2020arbert} are two BERT-based models, with the first focused on MSA ($61$GB) and the second on both MSA and dialects ($128$GB). MARBERT achieves SOTA on most Arabic NLU tasks. 
\noindent\textit{\textbf{QARiB}}~\cite{abdelali2021pre} is similarly a BERT-based model covering both MSA and dialects. \noindent\textit{\textbf{CamelBERT}}~\cite{CAMeLBERT2021} is also a BERT-based model pre-trained with MSA, dialectal, and classical Arabic. 

\label{sec:relwork}

\section{Conclusion}\label{sec:conclusion}
We introduced three powerful Arabic-specific text-to-text Transformer models trained on large MSA and/or Arabic dialectal data. We also introduced ARGEN, a unified benchmark for Arabic Natural Language \textit{generation} evaluation composed of \textit{seven} tasks collected from a total of $19$ datasets. 
Our models outperform  mT5 on \textit{all} ARGEN tasks ($52$ out of $59$ test sets, i.e., $88.14\%$). This is true even for MT involving four foreign languages from which the models have seen marginal or no pre-training data (i.e., zero- and few-shot pre-training). Our models also set new SOTA on the large Arabic language \textit{understanding} evaluation benchmark ARLUE. Our models involve vocabulary from $11$ languages other than Arabic, and hence can easily be further pre-trained/fine-tuned in these languages. Our models are publicly available, and ARGEN datasets are accessible from our repository.

\section*{Acknowledgements}\label{sec:acknow}
We gratefully acknowledge support from the Natural Sciences and Engineering Research Council of Canada (NSERC; RGPIN-2018-04267), the Social Sciences and Humanities Research Council of Canada (SSHRC; 435-2018-0576; 895-2020-1004), Canadian Foundation for Innovation (CFI; 37771), Compute Canada (CC),\footnote{\href{https://www.computecanada.ca}{https://www.computecanada.ca}}, UBC ARC-Sockeye,\footnote{\href{https://arc.ubc.ca/ubc-arc-sockeye}{https://arc.ubc.ca/ubc-arc-sockeye}} and Advanced Micro Devices, Inc. (AMD). We thank the
Google TFRC program for providing us with free TPU access.\footnote{\href{https://sites.research.google/trc/about/}{https://sites.research.google/trc/about/}} Any opinions, conclusions or recommendations expressed in this material are those of the author(s) and do not necessarily reflect the views of NSERC, SSHRC, CFI, CC, ARC-Sockeye, AMD, or Google. 


\section*{Ethics Statement}\label{sec:ethics}
\textbf{Energy efficiency.} Our models, similar to many deep learning language models, take significant pre-training time and are not energy efficient. We acknowledge this important issue and believe work on creating energy efficient models should receive scholarly attention. 

\noindent\textbf{Data.} Our pre-training datasets are collected from the public domain and cover diverse communities. As we have demonstrated, our resulting models are better equipped to power applications involving several varieties of Arabic as well as code-switched language use involving Arabic. From this perspective, we hope they add to ongoing efforts in the community to design models that are fairer and more representative.  

\noindent\textbf{ARGEN Benchmark Release.} We design ARGEN using both existing datasets and new datasets that we create for this work. In our accompanying GitHub repository, we link to all existing publicly available components of the benchmark with standard splits from source as well as components that can be acquired from data organizations. In addition, we released all the new datasets we have developed. While we have prioritized standardizing evaluation on as many unified and consolidated datasets and tasks as possible, we also report performance on individual test sets so as to enable the community to replicate our work even on particular parts or tasks of ARGEN if they so wish.  

\noindent\textbf{AraT5 Models Release.} All our pre-trained models are publicly available for non-malicious use. We acknowledge our models may still be misused in real world. However, we hope the models will be deployed in domains such as education, disaster management, health, recreation, travel, etc. in socially beneficial ways. These meaningful potential use cases are behind our decision to release the models.

\label{sec:conclusion}

\normalem
\bibliography{anthology,custom}
\bibliographystyle{acl_natbib}

\appendix
\clearpage
\twocolumn[{%
 \centering
 
}]

\appendixpage
\addappheadtotoc
\numberwithin{figure}{section}
\numberwithin{table}{section}





\section{A Study of Arabic mC4 Data Quality}\label{sec:ara-mc4-study}
\newcite{xue2020mt5} train mT5 on the mC4 dataset. They report $57$B Arabic tokens (almost double our token size) from $53$M webpages, making $1.66\%$ of all mT5 data. For our analysis, we randomly sample $1$M paragraphs from the Arabic part of mC4. We use paragraphs rather than whole documents for a more fine-grained analysis that is more comparable to our own data (especially in the case of Twitter). We first perform language identification using CLD3~\cite{mccandless2010accuracy} on the data. We find a sizable amount of the data (i.e., $13.59\%$) to be non-Arabic (mostly English or French). We manually inspect $\sim 100$ random samples of the data predicted as non-Arabic. We find these are mostly either non-linguistic content (e.g., java-script or HTML code) or non-Arabic text. The non-Arabic text is sometimes foreign language advertising or even full translation of the Arabic text in some cases. In many cases, non-Arabic is also boilerplate text such as that in web fora. Also, no samples of the non-Arabic included real \textbf{code-switching}. 

We also run an in-house MSA-dialect classifier on the same $1$M data sample. The classifier predicts an overriding majority of the data ($99.83\%$) as MSA. We again manually inspect $\sim 100$ samples from the small fraction predicted as dialects (i.e., $0.17\%$). While we find some of these to be actual dialectal text (usually short belonging to either Egyptian or Saudi dialects) from web fora, in the majority of cases the text is simply names of soap operas or advertisements. Our own pre-training data in the case of Twitter, in comparison, involve much more dialectal content ($28.39\%$ as listed in \cref{subsec:tr_data}). 



\section{ Evaluation on Arabic NLU}\label{sec:arbench} 
\subsection{ARLUE Benchmark}\label{subsec:arbench_cats}

Recently,~\newcite{abdul2020arbert} introduced ARLUE,  a natural language understanding benchmark for Arabic. ARLUE is composed of $42$  publicly available datasets, making it the largest and most diverse Arabic NLP benchmark.  ARLUE  is arranged into the six cluster tasks of sentiment analysis (SA), social meaning (SM), topic classification (TC), dialect identification (DI), named entity recognition (NER), and question answering (QA). We methodically evaluate each cluster task, ultimately reporting a single ARLUE score following~\newcite{abdul2020arbert}. Table~\ref{tab:Arbech_data}, shows a summary of the ARLUE benchmark. We briefly describe ARLUE tasks next.




\noindent \textbf{ARLUE\textsubscript{Senti}.}  To construct this task cluster~\newcite{abdul2020arbert}  merged  $17$ MSA and DA publicly available datasets. 

\noindent \textbf{ARLUE\textsubscript{SM}.} ARLUE\textsubscript{SM} refers to  eight social meaning datasets covering prediction of age, dangerous speech, emotion, gender, hate speech, irony, offensive language, and sarcasm. used in this benchmark. We will follow~\newcite{abdul2020arbert} in not merging the social meaning datasets, but rather report performance on each individual dataset as well as average performance across all tasks as part of an overall ARLUE score.

\noindent\textbf{ARLUE\textsubscript{Topic.}} This benchmark component  is a concatenation~\footnote{We note that the classes were straightforwardly merged without modifying any class labels.} of three topic classification datasets: Arabic News Text (ANT)~\cite{chouigui2017ant}, Khaleej~\cite{abbas2011evaluation}, and OSAC~\cite{saad2010osac}. %


\noindent\textbf{ARLUE\textsubscript{Dia.}}
Five datasets are used for dialect classification. These are AOC~\newcite{zaidan2014arabic}, ArSarcasm\textsubscript{Dia}~\cite{farha2020arabic},  MADAR (sub-task 2)~\cite{bouamor2019madar}, NADI-2020~\cite{mageed-etal-2020-nadi}, and QADI~\cite{abdelali2020arabic}. 

\noindent ARLUE\textsubscript{Dia} involve three  categories, namely, {\textbf{ARLUE\textsubscript{Dia-B}}} for MSA-dialect  classification (\textit{binary}). {\textbf{ARLUE\textsubscript{Dia-R}}}, and {\textbf{ARLUE\textsubscript{Dia-C}}}  for the region and country  level classification into four classes (\textit{region}), and $21$ classes (\textit{country}) respectively. 

 \noindent\textbf{ARLUE\textsubscript{QA}}. Four Arabic and multilingual  QA datasets are concatenated to build ARLUE\textsubscript{QA}:   ARCD \cite{mozannar2019neural} MLQA~\cite{lewis2019mlqa}, XQuAD~\cite{artetxe2020cross}, and TyDi QA~\cite{artetxe2020cross}.\footnote{All corresponding splits from the different QA datasets are merged.}

\subsection{ARLUE Evaluation}\label{subsec:LU_eval}

\textbf{Baselines.}
For comparison, we fine-tune a number of models on the same training data as our new models. These include the multilingual sequence-to-sequence model mT5~\cite{xue2020mt5}, and the  powerful  Arabic-specific BERT-based model MARBERT \cite{abdul2020arbert}. We note that MARBERT achieves the SOTA~\footnote{MARBERT outperform both multilingual encoder-only Transformers mBERT, XLM-R\textsubscript{Base},  XLM-R\textsubscript{Large}, and Arabic-specific BERT-based AraBERT \cite{antoun2020arabert}, ARBERT~\cite{abdul2020arbert}.} across the majority of  6 cluster tasks of ARLUE, with  the highest ARLUE score.

\noindent \textbf{Settings and Evaluation.}
We evaluate our models on the language understanding benchmark, ARLUE, under two settings: (i) single task learning and (ii) multi-task learning.  We present results on all the task clusters included in ARLUE except for NER which is a token-level task that is not straightforward with the text-to-text set up we adopt. Table~\ref{tab:ArBench_res_test} shows our evaluation results using the relevant metric for each task. 

\noindent\newcite{abdul2020arbert} introduced \textbf{ARLUE score}, a metric used to score pre-trained language model performance on multiple datasets. ARLUE score is a simply macro-average of the different scores across all task clusters, where  each task is weighted equally following~\cite{wang2018glue}. We compute the ARLUE score (i.e., overall macro-average) for each of our three models (i.e., AraT5\textsubscript{MSA}, AraT5\textsubscript{Tw}, and AraT5) and the baseline (mT5). 

\begin{table}[ht]
\centering
\resizebox{0.95\columnwidth}{!}{%
\begin{tabular}{lccrrr}
\hline
\textbf{Dataset} & \textbf{\#Datasets}& \textbf{Task} &\textbf{TRAIN}   & \textbf{DEV} & \textbf{TEST}             \\ \hline 
ARLUE\textsubscript{Senti} & $17$& SA & $190.9$K & $6.5$K  & $44.2$K  \\  

ARLUE\textsubscript{SM} &  $8$&SM & $1.51$M  &  $162.5$K  & $166.1$K  \\
ARLUE\textsubscript{Topic} &  $5$&TC & $47.5K$ & $5.9$K  & $5.9$K  \\  
ARLUE\textsubscript{Dia-B} &  $2$&DI & $94.9$K & $10.8$K  & $12.9$K  \\  
ARLUE\textsubscript{Dia-R} &   $2$& DI& $38.5$K & $4.5$K  & $5.3$K   \\  
ARLUE\textsubscript{Dia-C} &  $3$&DI & $711.9$K & $31.5$K  & $52.1$K  \\

ARLUE\textsubscript{QA}\textsuperscript{\textit{$\ddagger$}} & $4$& QA & $101.6$K  &  $517$  & $7.45$K  \\

\toprule
\end{tabular}%
}
\caption{\small{ARLUE categories across the different data splits. $^{\ddagger}$~Number of question-answer pairs~\cite{abdul2020arbert}. }  }
\label{tab:Arbech_data}
\end{table}
\begin{table}[ht]
\centering
 \renewcommand{\arraystretch}{1.3}
\resizebox{1\columnwidth}{!}{%
\begin{tabular}{lccccc}
\toprule 
\textbf{Dataset}     & \textbf{SOTA} & \textbf{mT5} & \textbf{AraT5\textsubscript{Tweet}} & \textbf{AraT5\textsubscript{MSA}} & \textbf{AraT5\textsubscript{}} \\
\toprule 
ARLUE\textsubscript{Senti}$^\star$                                           &   
$93.30$ / $94.00$	&	$92.46$ / $93.50$	&	$92.79$ / $93.50$	&	$\bf93.44$ / $\bf94.00$	&	$93.30$ / $94.00$                                 \\
ARLUE\textsubscript{SM}\textsuperscript{$\dagger$}  &
$81.60$ /$76.34$	&	$80.26$ / $73.59$	&	$80.41$ / $75.08$	&	$\bf 81.97$ / $\bf 76.60$	&	$81.09$ / $75.99$
        \\
ARLUE\textsubscript{Topic}                                           &     $90.07$ / $91.54$	&	$91.92$ / $93.36$	&	$90.86$ / $92.08$	&	 $\bf92.32$ / 93.30	&$\bf92.32$ / $\bf93.66$                                 \\
ARLUE\textsubscript{Dia-B}                                         &    $88.47$ / $87.87$	&	$86.48$ / $85.72$	&	$87.72$ / $87.06$	&	$\bf 88.51$ / $\bf 87.90$	&	$88.01$ / $87.41$                                  \\
ARLUE\textsubscript{Dia-R}                                         &  $90.04$ / $89.67$	&	$88.30$ / $87.93$	&	$90.12$ / $89.65$	&	$\bf91.17$ / $90.80$	&	$91.13$ / $\bf90.87$                                         \\
ARLUE\textsubscript{Dia-C}                                        &    $47.49$ / $38.53$	&	$45.94$ / $38.14$	&	$53.34$ / $42.02$	&	$52.65$ / $42.42$	&	$\bf53.64$ / $\bf43.18$                 \\

ARLUE\textsubscript{QA}$^\ddagger$                                             & $\bf40.47$ / $\bf62.09$	&	$36.92$ / $56.17$	&	$30.42$ / $49.57$	&	$39.47$ / $60.51$	&	$39.80$ / $60.93$\\    
\hline
Average                                                &     $75.92$ / $77.15$	&	$74.61$ / $75.49$	&	$75.09$ / $75.56$	&	$\bf77.08$ / $77.93$	&	$77.04$ / $\bf78.01$                 \\\toprule 
\textbf{ARLUE\textsubscript{Score}}                     & $76.53$	&	$75.05$	&	$75.33$	&	$77.50$	&	$\bf77.52$		 
\\    
\toprule             
\end{tabular}%
}
\caption{Performance of our models on ARLUE TEST datasets (Acc / F$_1$). $^\star$ Metric for ARLUE\textsubscript{Senti} is Acc/ F\textsubscript{1}\textsuperscript{PN}. $^\ddagger$  Metric for ARLUE\textsubscript{QA} is Exact Match (EM) / F$_1$.$^\dagger$  ARLUE\textsubscript{SM}  results is the average score across the social meaning tasks. \textbf{SOTA:}  MARBERT~\cite{abdul2020arbert}. }

\label{tab:ArBench_res_test}
\end{table}

\noindent\textbf{Single Task.} We fine-tune our three models and mT5 individually on each of the six tasks of ARLUE. We typically (i.e., in \textit{all} our experiments) identify the best checkpoint for each model on the development set, and report its performance on both development and test data. As Table~\ref{tab:ArBench_res_test} shows, our AraT5 model achieves the highest ARLUE score ($77.52$), followed by   AraT5\textsubscript{MSA}  ($77.50$) and  AraT5\textsubscript{TW} ($75.33$). We note that all our models outperform mT5 and the MARBERT (SOTA) by $\sim+2.74$ and $\sim+1$  ARLUE score points, respectively. 
\begin{table}[ht]
\centering
 \renewcommand{\arraystretch}{1.5}
\resizebox{0.95\columnwidth}{!}{%
\begin{tabular}{lccccc}
\toprule 
\textbf{Dataset}     & \textbf{S/M} & \textbf{mT5} & \textbf{AraT5\textsubscript{Tw}} & \textbf{AraT5\textsubscript{MSA}} & \textbf{AraT5\textsubscript{}} \\
\toprule 
\multicolumn{1}{c}{\multirow{2}{*}{ARLUE\textsubscript{Dia-B}}}
   & S                              & $86.48$ / $85.72$	&	$87.72$ / $87.06$	&	$\textbf{88.51}$ / \textbf{87.90}	&	$88.01$ / $87.41$                                     \\ 
   & M                                & $86.30$ / $85.54$	& $87.77$ / 	$87.20$                           & $87.93$ / $87.36$ & $88.02$ / $87.40$                                   \\\hline
\multicolumn{1}{c}{\multirow{2}{*}{ARLUE\textsubscript{Dia-R}}}                                         & S                                 & 	$88.30$ / $87.93$	&	$90.12$ / $89.65$	&	$91.17$ / $90.80$	&	$91.13$ / $90.87$                                          \\
& M                               &$89.01$ / $88.15$	& $91.53$ / $91.17$                           & $91.42$ / $91.15$ & $\textbf{91.51}$ / $\textbf{91.24} $                                  \\\hline
\multicolumn{1}{c}{\multirow{2}{*}{ARLUE\textsubscript{Dia-C}}}                                       & S                                & $45.94$ / $38.14$	&	$53.34$ / $42.02$	&	$52.65$ / $42.42$	&	$53.64$ / $43.18$                            \\
   & M                                & 45.86 / 38.12	& 53.42 / 40.86                           & 53.34 / 43.03 & \textbf{53.70}\textbf{} /\textbf{ 43.37}                                     \\
    \toprule             
\end{tabular}%
}
\caption{Performance of our models on ARLUE Dialects Test datasets on single and multi tasks setting (Acc / F$_1$). We copied single tasks results from Table~\ref{tab:ArBench_res_test} in this table for comparison.}  \label{tab:Arbechgen_multi_da}
\end{table}

\begin{table}[ht]
\centering
 \renewcommand{\arraystretch}{1.3}
\resizebox{0.95\columnwidth}{!}{%
\begin{tabular}{lccccc}
\toprule 
\textbf{Dataset}     & \textbf{S/M} & \textbf{mT5} & \textbf{AraT5\textsubscript{Tw}} & \textbf{AraT5\textsubscript{MSA}} & \textbf{AraT5} \\
\toprule 
\multicolumn{1}{l}{\multirow{2}{*}{Age}}
  & S                                  & $60.86$ / $61.05$	&	$62.29$ / $62.48$	&	$63.26$ / $63.41$	&	$63.50$ / $63.66$                                     \\ 
  & M                                 & $61.37$ / $61.47$	&	$63.92$ / $64.10$	&	$63.84$ / $38.41$	&	$\bf63.82$ / $\bf63.93$                                   \\\hline
   
  \multicolumn{1}{l}{\multirow{2}{*}{Dangerous}}
  & S                                  & $81.75$	/ $64.52$	&	$77.68 $/ $63.52$	&	$82.50$ / $66.93$	&	$75.41$ / $62.41$                                     \\ 
  & M                                 & $79.03$ / $66.46$	&	$\bf84.92$ / $68.73$	&	$84.46$ / $\bf71.62$	&	$77.53$ / $66.53$                                    \\\hline

\multicolumn{1}{l}{\multirow{2}{*}{Emotion}}
  & S                                  & $72.90$ / $71.34$	&	$73.65$ / $72.19$	&	$74.92$ / $73.30$	&	$\bf76.51$ / $\bf75.24$                                     \\ 
  & M                                 & $70.88$ / $68.87$	&	$72.79$ / $71.24$	&	$74.39$ / $73.08$	&	$74.28$ / $72.57$                                   \\\hline
\multicolumn{1}{l}{\multirow{2}{*}{Gender}}
  & S                                  & $72.05$ / $71.83$	& $72.27$ / $72.06$	& $73.83$ / $73.56$	& $73.38$ / $73.24$                                     \\ 
  & M                                 & $72.72$ / $72.42$	&	$74.58$ / $74.39$	&	$74.33$ / $74.23$	&	$\bf74.65$ / $\bf74.52$                                    \\\hline
\multicolumn{1}{l}{\multirow{2}{*}{Hate}}
  & S                                  & $95.70$ / $78.96$	&	$96.45$ / $81.75$	&	$\bf96.95$ / $\bf84.88$	&	$96.55$ / $83.33$                                     \\ 
  & M                                 & $95.75$ / $79.29$	&	$97.00$ / $82.73$	&	$96.40$ / $82.07$	&	$96.15$ / $80.39$                                    \\\hline    
\multicolumn{1}{l}{\multirow{2}{*}{Irony}}
  & S                                  & $82.61$ / $82.40$	&	$82.48$ / $82.25$	&	$83.23$ / $83.05$	&	$\bf82.98$ / $\bf82.80$                                   \\ 
  & M                                 & $80.99$ / $80.78$	&	$82.86$ / $82.65$	&	$82.86$ / $82.66$	&	$82.36$ / $82.21$                                    \\\hline
\multicolumn{1}{l}{\multirow{2}{*}{Offensive}}
  & S                                  & $91.35$ / $85.93$	&	$94.40$ / $90.96$	&	$94.15$ / $91.10$	&	$93.80$ / $90.11$                                     \\ 
  & M                                 & $90.30$ / $85.15$	&	$93.70$ / $90.41$	&	$\bf94.10$ / $90.83$	&	$94.05$ / $\bf90.85$                                    \\\hline

\multicolumn{1}{l}{\multirow{2}{*}{Sarcasm}}
  & S                                  & $84.83$ / $72.66$	&	$84.08$ / $75.42$	&	$86.92$ / $76.53$	&	$\bf86.59$ / $\bf77.13$                                     \\ 
  & M                                 & $84.64$ / $74.06$	&	$85.55$ / $75.25$	&	$86.26$ / $77.06$	&	$86.26$ / $76.63$                                    \\\toprule 
\textbf{\multirow{2}{*}{ARLUE\textsubscript{SM} }}                                          
&    S                          	&	$80.26$ / $73.59$	&	$80.41$ / $75.08$	&	$81.97$ / $\bf76.60$	&	$81.09$ / $75.99$                             \\
& M                                 & $79.46$ / $73.56$	&	$81.92$ / $76.19$	&	$\bf82.08$ / $73.75$	&	$81.14$ / $75.95$ \\
\toprule             
\end{tabular}%
}
\caption{Performance of our models on ARLUE social meaning (SM) Test datasets on single- and multi-tasks setting (Acc / F$_1$). \textbf{S}: Single Task.\textbf{ M}:Multi-task. }\label{tab:Arbechgen_multi_sm} 
\end{table}

\noindent\textbf{Multitask.} We also investigate multitask learning \cite{caruana1997multitask,ruder2017overview} with our AraT5 models. This approach consists of training the model on multiple tasks simultaneously (i.e., the model and its parameters are shared across all tasks) in order to eventually improve performance on each individual task. In our case, we fine-tune our models on many tasks at the same time using: (i) The three dialect datasets: ARLUE\textsubscript{Dia-B},  ARLUE\textsubscript{Dia-R}, and  ARLUE\textsubscript{Dia-C} and (ii) the social meaning datasets of ARLUE\textsubscript{SM}.  Table~\ref{tab:Arbechgen_multi_da} and  Table~\ref{tab:Arbechgen_multi_sm} show the results of  multi-task  experiments for dialect settings and social meaning, respectively. Our results show that multi-task training outperforms single task models in the majority of the dialects experiments (n=$7$ out of $9$ experiments, ~$77.78$\% of the tasks) and half of the social meaning tasks (n=$18$ out of $36$ experiments, $50$\% of the tasks). These results are promising, and hence we plan to further investigate multi-task learning with our new models in the future.





\section{ARGEN}\label{sec:ARGEN_app}

\subsection{Arabic Paraphrase Data}\label{sec:ARGEN_parah_app}
\textbf{AraPara.} is a new multi-domain Arabic paraphrasing dataset we create using English-Arabic parallel OPUS data~\cite{OPUS}. To ensure high-quality, we follow four careful steps: \textbf{(1)} We pick $1$ million English-Arabic parallel sentences from  OPUS~\cite{OPUS} covering the different domains. \textbf{(2)} We translate the  English  sentences using a high-quality in-house English$\rightarrow$Arabic MT model. \textbf{(3)} We run the multi-lingual semantic similarity model from~\newcite{yang2019multilingual} on the Arabic machine translated sentences and the human translation (i.e., original Arabic sentences from OPUS), keeping only sentences with an arbitrary semantic similarity score between $0.70$ and $0.99$. This allows us to filter out identical sentence pairs (i.e., similarity score = $1$) and those that are not good translations (i.e., those with a  semantic similarity score $< 0.70$). \textbf{(4)} In order  to  maximize syntactic and lexical diversity of the pairs of paraphrased sentences, we perform an analysis based on word overlap between the semantically similar pair sentences (i.e., the output  of the previous step). We then perform a \textit{manual} analysis of the data, identifying sentences with unigram token overlap between $35\%$ and $70\%$ as sufficiently distinct paraphrase pairs. This gives us $122$K paraphrase pairs. We split these sentence pairs into $116$K for training and $6$K for validation. 

\subsection{Evaluation on DEV}\label{sec:Arabic_LG_eval_app}

\begin{table}[]
 \centering
 \footnotesize
 \renewcommand{\arraystretch}{1.2}
\resizebox{0.9\columnwidth}{!}{%
 \begin{tabular}{lccc}
 \toprule
\textbf{Split} & \textbf{Article/Title}   &  \textbf{Avg article len} & \textbf{Avg title len}\\
\toprule
\textbf{TRAIN}     &    $93.3$K & $256.46$   &$10.06$  \\
\textbf{DEV}        &   $11.7$K  & $253.11$    & $10.03$  \\
\textbf{TEST}        &  $11.7$K & $260.32$   & $10.03$   \\
\cline{1-4}
\textbf{Total}        &  $116.6$K & $256.63$   &$10.04$   \\
\toprule 
\end{tabular}%
}
\caption{\footnotesize Main characteristics of  ARGEN\textsubscript{NTG} data splits. For each split, we provide the number of article-title pairs and the average length of the articles and titles.  }  \label{tab:arnews_tg}

    \end{table}
\begin{table*}[]
\centering
 \renewcommand{\arraystretch}{1.2}
\resizebox{.9\textwidth}{!}{%
\begin{tabular}{lllllcc}
\toprule
 \textbf{Varieties}   & \textbf{Dataset}        & \textbf{Region}     &  \textbf{Country-Level}  & \textbf{City-Level}           & \textbf{DEV}            & \textbf{TEST}                  \\
\toprule

&\multirow{2}{*}{{ADPT}~\newcite{zbib2012machine} }   &  Levantine  & -   &  -  & -   &  $138$K    \\
 & &  Nile   &Egypt &  - & -  & $38$K                   \\   \cdashline{2-7}
&\multirow{2}{*}{{Bible I} }  & \multirow{2}{*}{Maghrebi}     &Tunisia &  -  &   -   &   $600$                   \\
 & &    & Morocco   &  -  &  -   &   $600$                 \\ \cdashline{2-7}

\multirow{18}{*}{{\textbf{DIA}}}  &\multirow{24}{*}{{MADAR I}~\newcite{bouamor2018madar}  }  & \multirow{4}{*}{Nile}   & Egypt   & Cairo   &      -   &  $6.5$k           \\
& &   & Egypt  & Alexandria   &     -   &   $2$k           \\
& &   &  Egypt  & Aswan   &     -   &   $2$k           \\
& &   & Sudan  & Khartoum  &     -   &   $2$k           \\ \cdashline{3-7}
& & \multirow{8}{*}{{Gulf}}   &  Qatar & Doha   &      -   &  $6.5$k           \\
& &     &  Yemen &  Sana'a   &   -   &     $2$k           \\
& &  &  Oman & Muscat   &    -   &    $2$k           \\
& &  &  KSA & Riyadh   &     -   &   $2$k          \\
& &  &  Jedd & Muscat   &     -   &   $2$k          \\
& &  &  Iraq & Baghdad   &    -   &   $2$k           \\ 
& &  &  Iraq & Basra   &     -   &   $2$k           \\ 
& &  &  Iraq & Mosu   &     -   &   $2$k          \\  \cdashline{3-7}

& &\multirow{6}{*}{{Leventian}}  &  Lebanon & Beirut   &    -   &     $6.5$k         \\
& &  &  Palestine  & Jerusalem   &      -   &  $2$k       \\
& &  &  Jordan & Amman   &     -   &   $2$k           \\
& &  &  Jordan & Salt.   &      -   &  $2$k           \\
& &  &  Syria & damascus   &    -   &    $2$k           \\
& &  &  Syria & Alep   &      -   &  $2$k           \\  \cdashline{3-7}
& &\multirow{6}{*}{{Maghrebi}}    &  Algeria  & Alger   &      -   &  $2$k       \\ 
& &   &  Lybia  & Trip   &       -   & $2$k      \\ 
& &   &  Lybia  & Beng   &       -   & $2$k       \\ 
& &   &  Tunisia  & Tunis   &    -   &    $6.5$k       \\ 
& &   &  Tunisia  & Safax   &    -   &    $2$k       \\ 
& &   &  Morocco  & Fes   &      -   &   $6.5$k     \\ 
& &   &  Morocco  & Rabat   &     -   &   $2$k       \\ 
\toprule
\multirow{6}{*}{{\textbf{MSA}}}  & \multirow{2}{*}{{Bible II}}  &  -    & -  &  -   & -  & $600$                   \\
  &  &  -    & -  &  - & -  & $600$                   \\
  \cdashline{2-7}
   & {MADAR II}~\newcite{bouamor2018madar}     & -   &  - & -   & -   &     $6.5$k       \\ 
   \cdashline{2-7}
   & {IWSLT TED15}~\newcite{cettolo2016iwslt} &  -  & -  & -  & -  & $1.1$k                  \\
    & {IWSLT TED16 / }~\newcite{cettolo2016iwslt} &  -  & -  & -  & -  & $1.1$k                    \\
   & {IWSLT QED16}~\cite{cettolo2016iwslt} &-&  -  & - & -  & $550$                   \\ 
   \cdashline{2-7}
    &  {UN} \newcite{ziemski2016united} & - & - & - & $4$k &  $4$k            \\       
        &  {OPUS-X-Ara}  & - & - & - & $5$k &  $5$k                
    \\



\toprule
\end{tabular}}
\caption{Arabic to English datasets included in ARGEN\textsubscript{MT}.  \textbf{MADAR~I:} corpus consists of 2k sentences (Test) of 21 city-level dialects each.  \textbf{MADAR~II:}  12k sentences (5.5k  for Dev, and 6.5k  for Test sets) each of five other city-level dialects and MSA. \textbf{Bible~I:} $600$ sentences each as Dev and Test sets for Moroccan, Tunisian,  and MSA.\textbf{ Bible~II:} Two  Dev and Test splits ($600$ sentences each) are used for Bible MSA. }
\label{tab:mt_data_app}

\end{table*}
In this section we describe the ARGEN\textsubscript{MT} datasets splits and report the evaluation results in validation datasets. Details about ARGEN\textsubscript{NTG} are in Table~\ref{tab:arnews_tg} and ARGEN\textsubscript{MT} datasets splits are shown in Table~\ref{tab:mt_data_app}. Moreover, The evaluation on validation datasets for ARGEN\textsubscript{TS} are described in Table~\ref{tab:rs_mt_data_dev} and \ref{tab:Arbechgen_summary_DEV}, respectively. Finally, Table~\ref{tab:NGT_QA_SEV} shows the validation results of ARGEN\textsubscript{NTG}, ARGEN\textsubscript{QG}, ARGEN\textsubscript{TR}, and ARGEN\textsubscript{PHP} datasets.

\begin{table*}[]
\centering
\footnotesize
 \renewcommand{\arraystretch}{1.1}
\resizebox{.9\textwidth}{!}{
\begin{tabular}{lllHrrrrrr|r}
\toprule
\multicolumn{2}{c}{ \textbf{\footnotesize Dataset}   }    & \textbf{\footnotesize Test Split}  & \textbf{\footnotesize SOTA}    & \textbf{S2S\textsubscript{2M}}  & \textbf{S2S\textsubscript{10M}}&  \textbf{ \footnotesize mT5}  & \textbf{\footnotesize AraT5\textsubscript{Tw}} & \textbf{ \footnotesize AraT5\textsubscript{MSA}} & \textbf{ \footnotesize AraT5\textsubscript{}} & \textbf{\footnotesize SOTA} \\
\toprule

&\multirow{2}{*}{{\textbf{ADPT}}\textsuperscript{$\dagger$} }   &  Lev &  $\bf11.00$&$4.90$&$7.50$&$10.12$&$\bf10.53$&$9.33$&$9.53$ &  $11.00$    \\
&    &  Egy &  $\bf13.40$&$5.04$&$9.21$&$11.63$&$10.68$&$11.33$&$\bf11.87$& $13.40$    \\
\cdashline{2-11}
&\multirow{2}{*}{{\textbf{Bible I}\textsuperscript{$\dagger$}} }  & \multirow{1}{*}{Tun.}     & $\bf7.20$&$4.44$&$4.80$&$6.98$&$4.63$&$\bf7.48$&$6.50$    & $7.20$      \\
 & &     \multirow{1}{*}{Mor.}      &$\bf4.10$&$3.22$&$3.47$&$\bf7.65$&$5.98$&$8.25$&$7.83$    &$4.10$        \\ \cdashline{2-11}

\multirow{6}{*}{{\textbf{DA}}}& \multirow{5}{*}{\textbf{MADAR I}\textsuperscript{$\dagger$}}  & \multirow{1}{*}{Egy.} &$\bf27.1$&$17.1$&$17.71$&$24.07$&$21.68$&$\bf24.75$&$24.29$  &$27.1$  \\
& & \multirow{1}{*}{{Qat.}}                                   &$\bf28.10$&$16.52$&$17.92$&$23.45$&$22.32$&$\bf23.98$&$23.58$  &$28.10$ \\  
& &\multirow{1}{*}{{Leb.}}                                      & $\bf21.80$&$9.61$&$12.93$&$18.19$&$16.06$&$\bf18.64$&$16.82$ & $21.80$\\ 
& &\multirow{1}{*}{{Tun.}}                                        & $\bf12.10$&$9.06$&$9.30$&$10.62$&$9.23$&$\bf10.97$&$10.25$   & $12.10$  \\  
& &\multirow{1}{*}{{Mor.}}                                        & $\bf10.00$&$8.46$&$8.40$&$11.83$&$8.39$&$\bf12.09$&$11.26$   & $10.00$\\   \cdashline{2-11}
&\textbf{QAraC}\textsuperscript{$\dagger$} & $-$ & $\bf11.70$&$10.31$&$10.46$&$11.87$&$10.73$&$\bf11.30$&$10.64$  & $11.70$ \\

\toprule
\multirow{6}{*}{{\textbf{MSA}}}  & \multirow{2}{*}{{\textbf{Bible II}\textsuperscript{$\dagger$}}}  &  Test 1   &$\bf16.60$&$11.43$&$11.33$&$15.68$&$13.13$&$\bf16.43$&$15.89$   &$16.60$  \\
  &   & Test 2   & $12.9$&$5.88$&$6.41$&$12.76$&$9.69$&$\bf13.53$&$11.96$  & $12.9$        \\ \cdashline{2-11}
  & \textbf{MADAR I}\textsuperscript{$\dagger$}   & MSA & $\bf45.8$&$40.75$&$41.84$&$39.11$&$38.06$&$\bf39.92$&$39.25$ & $45.8$     \\   \cdashline{2-11}
  &  \textbf{IWSLT}\textsuperscript{$\ddagger$}& QED16  &$--$&$28.39$&$29.04$&$29.18$&$28.59$&$\bf30.19$&$29.97$ &$-~$             \\ \cdashline{2-11}
   
    & \textbf{UN}\textsuperscript{$\dagger\dagger$} &Ar-En&$--$&$51.54$&$51.97$&$50.84$&$50.14$&$\bf52.11$&$51.54$ &$-~$              \\

\toprule
  &  &    \textbf{\textit{Average}} & $17.06$&$14.67$&$15.66$&$18.50$&$16.94$&$\bf18.90$&$18.31$& $17.06$\\


\toprule

\end{tabular}}
\caption{ARGEN\textsubscript{MT}  datasets on Dev splits. \textbf{S2S:} Sequence-to-sequence Transformer models trained from scratch without use of a language model. \textbf{SOTA:} \textsuperscript{$\dagger$}\cite{sajjad2020arabench}, \textsuperscript{$\ddagger$}\cite{durrani2017qcri}, \textsuperscript{$\dagger\dagger$}\cite{junczys2016neural}.}
\label{tab:rs_mt_data_dev}

\end{table*}

\begin{table}[]
\centering
 \renewcommand{\arraystretch}{1.2}
\resizebox{0.95\columnwidth}{!}{%
\begin{tabular}{lccccc}
\toprule 
\textbf{Dataset}     & \textbf{Metric} & \textbf{mT5} & \textbf{AraT5\textsubscript{Tweet}} & \textbf{AraT5\textsubscript{MSA}} & \textbf{AraT5\textsubscript{}} \\\toprule
\multicolumn{1}{c}{\multirow{3}{*}{WikiLin.}}
   & Rouge1                                  & $71.03$	&	$\bf74.20$ &	$72.64$ &	$73.87$                                    \\ 
   & Rouge2                                 &  $62.87$                      & $\bf66.37$ &	$64.24$	& $65.76$                                    \\
    & RougeL                                  & $70.99$	&	$\bf74.14$ &	$72.55$	&$73.79$                                    \\ 
   \hline
    
\toprule 
\end{tabular}%
}
\caption{Performance of our models on document summarization Dev splits. }  \label{tab:Arbechgen_summary_DEV}
\end{table}

\begin{table}[]
\centering
 \renewcommand{\arraystretch}{1.2}
\resizebox{0.95\columnwidth}{!}{%
\begin{tabular}{lHcccc}
\toprule 
\textbf{Dataset}     & \textbf{Metric} & \textbf{mT5} & \textbf{AraT5\textsubscript{Tweet}} & \textbf{AraT5\textsubscript{MSA}} & \textbf{AraT5} \\\toprule
ARGEN\textsubscript{NTG}
   
       & BLUE   &$19.22$	&$19.38$&	$\bf20.19$	&$20.01$          \\

ARGEN\textsubscript{QG}

           & BLUE                                  & $13.95$ &	$11.25$	&$12.96$	&$\bf15.36$          \\
           
ARGEN\textsubscript{TR} & BLUE                                  & $64.81$ &	$62.95$	&$\bf69.30$	&$65.54$          \\           
           
ARGEN\textsubscript{PHP} & BLUE                                  & $30.70$ &	$31.54$	&$\bf33.15$	&$32.36$          \\    
   \hline
\toprule

\end{tabular}%
}
\caption{Performance of our models on title, question generation, transliteration, and paraphrasing DEV split based on Bleu score.}  \label{tab:NGT_QA_SEV}
\end{table}

\section{Qualitative Analysis of Models }\label{sec:qualitative_analysis_app}
\begin{table}[t]
\centering
\resizebox{0.9\columnwidth}{!}{%
\begin{tabular}{lcccc}
\toprule
  \textbf{Dataset}  &  \textbf{mT5} & \textbf{AraT5\textsubscript{Tweet}} & \textbf{AraT5\textsubscript{MSA}} & \textbf{AraT5\textsubscript{}} \\ \hline
    & \multicolumn{4}{c}{All Length}                                                                                                                                                                                                        \\ \hline
MSA & \multicolumn{1}{c}{$28.38$}         & \multicolumn{1}{c|}{$27.03$}     & \multicolumn{1}{c|}{\textbf{$\bf29.16$}} & $28.65$                      \\ 
DA  & \multicolumn{1}{c}{$20.19$}         & \multicolumn{1}{c|}{$17.73$}     & \multicolumn{1}{c|}{\textbf{$\bf20.54$}} & $20.10$                       \\ 
All & \multicolumn{1}{c}{$21.14$}         & \multicolumn{1}{c|}{$18.83$}     & \multicolumn{1}{c|}{\textbf{$\bf21.55$}} & $21.09$                      \\ \hline
    & \multicolumn{4}{c}{Sequence  length $<$ $10$}                                                                                                                                                                                       \\ \hline
MSA & \multicolumn{1}{c}{$35.73$}         & \multicolumn{1}{c|}{$35.50$}      & \multicolumn{1}{c|}{\textbf{$\bf36.96$}} & $36.44$                      \\ 
DA  & \multicolumn{1}{c}{$20.81$}         & \multicolumn{1}{c|}{$18.73$}     & \multicolumn{1}{c|}{\textbf{$\bf21.29$}} & $20.68$                      \\ 
All & \multicolumn{1}{c}{$21.70$}          & \multicolumn{1}{c|}{$19.75$}     & \multicolumn{1}{c|}{\textbf{$\bf22.23$}} & $21.65$                      \\ \hline
    & \multicolumn{4}{c}{$20$ $\leq$ Sequence  length $\leq$ $10$}                                                                                                                                                                                      \\ \hline
MSA & \multicolumn{1}{c}{$26.18$}         & \multicolumn{1}{c|}{$24.31$}     & \multicolumn{1}{c|}{\textbf{$\bf26.90$}}  & $26.24$                      \\ 
DA  & \multicolumn{1}{c}{$19.74$}         & \multicolumn{1}{c|}{$16.30$}      & \multicolumn{1}{c|}{\textbf{$\bf19.78$}} & $19.56$                      \\ 
All & \multicolumn{1}{c}{$21.03$}         & \multicolumn{1}{c|}{$17.94$}     & \multicolumn{1}{c|}{\textbf{$\bf21.22$}} & $20.91$                      \\ \hline
    & \multicolumn{4}{c}{$20$ $<$ Sequence  length }                                                                                                                                                                                       \\ \hline
MSA & \multicolumn{1}{c}{\textbf{$\bf19.50$}} & \multicolumn{1}{c|}{$16.91$}     & \multicolumn{1}{c|}{$19.28$}          & $19.45$                      \\ 
DA  & \multicolumn{1}{c}{$13.51$}         & \multicolumn{1}{c|}{$11.52$}     & \multicolumn{1}{c|}{\textbf{$\bf13.69$}} & $13.44$                      \\ 
All & \multicolumn{1}{c}{$15.20$}          & \multicolumn{1}{c|}{$13.05$}     & \multicolumn{1}{c|}{\textbf{$\bf15.26$}} & $15.13$                      \\ \toprule
\end{tabular}%
\caption{Sequence  length  based results on  ARGEN\textsubscript{MT}  Test datasets. }
\label{tab:seq_length_test}
}
\end{table}

In this section, we explore ability of our models to generate MSA and dialectal Arabic under various conditions. We now overview various types of analyses in this regard. While samples presented here are handpicked, we note that they are mostly representative of outputs from our models since we mainly chose them to demonstrate different linguistic attributes that we believed would be relevant to the analysis.

\noindent\textbf{Effect of Sample Length on MT.} We were inquisitive how \textbf{MT models} fine-tuning our pre-trained language models compare to mT5 under \textbf{different length conditions}. For this, we \textbf{(1)} merge all MSA and dialectal Test datasets in our Arabic$\rightarrow$English experiments to form a single dataset that we then \textbf{(2)} split into three bins/Test sets based on sentence length as shown in Table~\ref{tab:seq_length_test}. As the Table shows, our AraT5\textsubscript{MSA} outperform mT5 in \textit{all} but one condition (where our model acquires marginally less performance). We also performed similar evaluation on the merged Dev sets of all MSA and dialectal Arabic MT datasets in the Arabic$\rightarrow$English direction. We do not show related results here, but we note our AraT5\textsubscript{MSA} outperforms mT5 on \textit{all} conditions.

\noindent\textbf{MT Model Output.} Table~\ref{tab:msa_dia_examples_MT} shows three examples of Arabic$\rightarrow$English MT models. Sentence (1) is in \textbf{MSA source}, sentence (2) is in Levantine Arabic source, and sentence (3) is in Egyptian source. In all three examples, on or more of our models generate(s) more fluent translations than mT5. This includes ability of our models to translate dialectal sentences where mT5 seems to struggle (e.g., mT5 is not able to translate the equivalents of ``drive" from Egyptian Arabic).

\noindent\textbf{Code-Switched Translation Model Output.} Table~\ref{tab:cs_examples_MT} shows two code-switched examples from \textbf{ARGEN\textsubscript{CS}}. Sentence (1) is Algerian dialect at source translated into French, while sentence (2) Jordanian dialect translated into English. In both cases, our models not only handle the dialects but also their use in code-switched contexts better than mT5.

\noindent\textbf{Paraphrasing, Transliteration, and Title Generation Output.} Tables~\ref{tab:php_examples},~\ref{tab:tr_examples}, and~\ref{tab:ngt_examples} each shows two output samples from our paraphrasing, transliteration, and title generation models, respectively. In each case, the samples are high-quality, informative, and fluent. Our paraphrase samples also tightly capture the meaning of the source sentences.

\begin{table*}[ht]
\centering
 \renewcommand{\arraystretch}{1.7}
\resizebox{\textwidth}{!}{%
\begin{tabular}{ll}
\toprule 



\textbf{(1) Source:} & \begin{tabular}[c]{@{}l@{}}
~~\<هل تعرفون أن أحد المتع الكبيرة للسفر وأحد مباهج أبحاث الإثنوجرافيا في فرصة العيش بين أولئك الذين لم ينسوا الأساليب في الرياح ويلمسونه  >    \textbf{:MSA}\\
~~~~~~~~~~~~~~~~~~~~~~~~~~~~~~~~~~~~~~~~~~~~~~~~~~~~~~~~~~~~~~~~~~~~~~~~~~~~~~~~~~~~~~~~~~~~~~~~~~~~~~~~~~~~~~~~~~~~~~~~~~~~~~~~~~~~~~~~~ \< في الأحجار  التي صقلتها الأمطار  ويتذوقونه في أوراق النباتات المرة  >\\

\end{tabular}                                                       \\\cline{2-2}

\textbf{Target:} &   \begin{tabular}[c]{@{}l@{}}
\textbf{EN:}  Do you know that one of the intense pleasures of travel and one of the delights of ethnographic research is the opportunity to live amongst those\\ who have not  forgotten the old ways, who still feel their past in the wind, touch it in stones polished by rain, taste it in the bitter leaves of plants.

\\  

\end{tabular}                                                       \\\cline{2-2}
\textbf{\textbf{mT5}}   & \begin{tabular}[c]{@{}l@{}}


\colorbox{red!10}{you know, one} \colorbox{green!10}{of the great \textbf{enjoyments} of travel and one of the pleasure ofs} \colorbox{red!10}{statistics} \colorbox{green!10}{research is the \textbf{opportunity} to live among those who } \\ \colorbox{green!10}{have not forgotten old methods, who still feel their past in} \colorbox{red!10}{wind, touch the rain-saving stones} \colorbox{green!10}{and taste it in the snail of plants.}

\end{tabular}\\

\cdashline{2-2}
\textbf{\textbf{AraT5\textsubscript{Tw}}}   &

\begin{tabular}[c]{@{}l@{}}

\colorbox{red!10}{you know, one} \colorbox{green!10}{of the big pleasures of travel and one of the} \colorbox{red!10}{physical} \colorbox{green!10}{research} \colorbox{red!10}{approaches is a living} \colorbox{green!10}{\textbf{chance} among those who have not} \\\colorbox{red!10}{forgetted} \colorbox{green!10}{old methods, who still feel their past in the wind, touch it in}\colorbox{red!10}{ the stones that rained}\colorbox{green!10}{ and taste it in the} \colorbox{red!10}{fresh} \colorbox{green!10}{plant leaves}.

\end{tabular}\\\cdashline{2-2}

\textbf{\textbf{AraT5\textsubscript{MSA}}}     & \begin{tabular}[c]{@{}l@{}}

\colorbox{green!10}{\textbf{Do you know that} one of the great \textbf{pleasures} of travel and one of the joys of \textbf{ethnographic} research is the \textbf{opportunity} to live among those who have } \\ \colorbox{green!10}{not forgotten the ancient methods, who still feel their past in the wind, touch it in} \colorbox{red!10}{rain-purified} \colorbox{green!10}{stones and taste it in the bitter leaves of plants}\colorbox{red!10}{?}

\end{tabular}  \\
 \cdashline{2-2}
\textbf{\textbf{AraT5}}   & 	\begin{tabular}[c]{@{}l@{}}

\colorbox{red!10}{you know, one} \colorbox{green!10}{of the great} \colorbox{red!10}{benefits}  \colorbox{green!10}{of travel and one of the} \colorbox{red!10}{physiology} \colorbox{green!10}{research is the opportunity to live among those who have not} \\ \colorbox{green!10}{ forgotten the old methods, who still feel their  past in the wind, they feel their past in the \textbf{ stones that are refined by rain}, and they taste it}\colorbox{red!10}{ in the leaf.}

\end{tabular} \\

\hline

\textbf{(2) Source:} &  ~~~~~~~~~~~~~
~~~~~~~~~~~~~~~~~~~~~~~~~~~~~~~~~~~~~~~~~~~~~~~~~~~~~~~~~~~~~~~~~~~~~~~~~~~~~~~~~~~~~~~~~~~~~~~~~~~~~ \<عمفتش على مطعم رايق و حلو للشوي . بتحط عليهن إشارة عهالخريطة؟  >\textbf{~~~:LEV}

\\\cline{2-2}

\textbf{Target:} & \textbf{EN:}   I'm looking for a nice, quiet grill-type restaurant. would you point them out on this map?  

\\\cline{2-2}
\textbf{mT5} & 
   
\colorbox{red!10}{You find} \colorbox{green!10}{a nice and sweet} \colorbox{red!10}{cooking} \colorbox{green!10}{restaurant with a map sign?} 
\\\cdashline{2-2}

\textbf{\textbf{AraT5\textsubscript{Tw}}}     & \colorbox{red!10}{a snack on } \colorbox{green!10}{a nice and sweet sweat restaurant} \colorbox{red!10}{snack} \colorbox{green!10}{, you put on them a map sign?  }                                                 \\\cdashline{2-2}
\textbf{\textbf{AraT5\textsubscript{MSA}}}  & \colorbox{red!10}{You're} \colorbox{green!10}{\textbf{looking} at a nice and sweet} \colorbox{red!10}{snack} \colorbox{green!10}{restaurant} \colorbox{red!10}{with a sign on the map?  }                                                                     
  \\\cdashline{2-2}
\textbf{\textbf{AraT5}}  & \colorbox{green!10}{looking for a nice and sweet restaurant} \colorbox{red!10}{to eat}, \colorbox{green!10}{put a sign on them} \colorbox{red!10}{for} \colorbox{green!10}{the map? }                                                                                        \\
\hline 
\textbf{(3) Source} &  ~~~~~~~~~~~~~~~~~~~~~~~~~~~~~~~~~~~~~~~~~~~~~~~~~~~~~~~~~~~~~~~~~~~~~~~~~~~~~~~~~~~~~~~~~~~~~~~~~~~~~~~~~~~~~~~~~~~~~~~~~~~~~~~~~~~~~~~~~~~~\begin{tabular}[c]{@{}l@{}}\<ده فعلا مختلف ان الواحد يسوق على جنب الشارع اليمين.>   \textbf{:EGY} \end{tabular}  \\
\cline{2-2}
\textbf{Target:} & \textbf{EN:}  It's really different driving on the right side of the street.   

\\\cline{2-2}
\textbf{mT5} & 
   
\colorbox{green!10}{that's} \colorbox{green!10}{really different that one } \colorbox{red!10}{walks} \colorbox{green!10}{on the right side of the street.}    
\\\cdashline{2-2}

\textbf{\textbf{AraT5\textsubscript{Tw}}}   & 
   
\colorbox{green!10}{that's} \colorbox{green!10}{really different that one \textbf{drives}} \colorbox{red!10}{by} \colorbox{green!10}{the right side of the street.}    
\\\cdashline{2-2}

\textbf{\textbf{AraT5\textsubscript{MSA}}}   & 
   
\colorbox{green!10}{That's} \colorbox{green!10}{really different that one \textbf{runs}} \colorbox{green!10}{on the right side of the street.}    
\\\cdashline{2-2}

\textbf{\textbf{AraT5}}   & 
   
\colorbox{green!10}{That's} \colorbox{green!10}{really different that one \textbf{drives}} \colorbox{green!10}{on the right side of the street.}    \\

\toprule 
\end{tabular}%
}
\caption{MSA and DIA sentences with their English translations using our Models and mT5. Data samples are extracted from the Dev datasets. \colorbox{green!10}{\textbf{Green}} refers to good translation. \colorbox{red!10}{\textbf{Red}} refers to problematic translation.}
    \label{tab:msa_dia_examples_MT}
\end{table*}
\clearpage
\begin{table*}[]
\centering
 \renewcommand{\arraystretch}{0.5}
\footnotesize
\resizebox{0.7\textwidth}{!}{%
\begin{tabular}{ll}
\toprule 

\textbf{(1) Source } & \small \colorbox{blue!10}{ \textbf{Arabic sent:} }  \<أولا ، تبسيط و توضيح هذه المعاهدات>   \\
\cline{2-2}
\textbf{Target} &   \small \colorbox{blue!10}{\textbf{Paraphrase:}  }
\small \<.	في البداية ، يجب تبسيط و توضيح هذه المعاهدات .>
\\
\cline{2-2}

\textbf{mT5} &   \small  \<أولا، تبسيط وتوضيح المعاهدات.>
\\

\cdashline{2-2}

\textbf{\textbf{AraT5\textsubscript{Tw}}}    &    \small
\<أولا ، تبسيط المعاهدات وتوضيحها.>
\\
\cdashline{2-2}

\textbf{\textbf{AraT5\textsubscript{MSA}}}    &  \small \<أولا، تبسيط و توضيح هذه المعاهدات.>                                    \\    \cdashline{2-2}

\textbf{\textbf{AraT5}}    &  \small  \<أولا ، تم تبسيط المعاهدات وتوضيحها.>  \\               \hline


\textbf{(2) Source } & \small \colorbox{blue!10}{\textbf{Arabic sent:}  } \<قد يواجه عقوبة الإعدام إذا ما تمت إدانته بتهمة التجسس .>  \\
\cline{2-2}
\textbf{Target} &   \small \colorbox{blue!10}{\textbf{Paraphrase:} } 
\small \<يمكن لتهم التجسس ومساعدة العدو أن تودي لعقوبة الإعدام.>
\\
\cline{2-2}

\textbf{mT5} &   \small  \<ويمكن أن يواجه عقوبة الإعدام إذا ادعى أن هناك جريمة تجسس.> \\

\cdashline{2-2}

\textbf{\textbf{AraT5\textsubscript{Tw}}}    &    \small 
\<يمكن أن يحكم عليه بالإعدام إذا أدين بالتجسس.>
\\
\cdashline{2-2}

\textbf{\textbf{AraT5\textsubscript{MSA}}}    &  \small \<وقد يعاقب بالإعدام إذا أدين بتهمة التجسس.>                    \\    \cdashline{2-2}

\textbf{\textbf{AraT5}}    &  \small  \<ويجوز الحكم عليه بالإعدام إذا أدين بالتجسس.>\\          

\toprule

\end{tabular}%
}
\caption{ \colorbox{blue!13}{Paraphrasing} examples  extracted from Dev data splits.}
    \label{tab:php_examples}
\end{table*}

\begin{table*}[]
\centering
 \renewcommand{\arraystretch}{0.5}
\footnotesize
\resizebox{.7\textwidth}{!}{%
\begin{tabular}{ll}
\toprule 
\textbf{(1) Source } &  \small \textbf{\colorbox{green!10}{Arabizi:}}   Tab matsha3'ali 5edmt el iphone men Vodafone \\
\cline{2-2}
\textbf{Target} &  \small \textbf{\colorbox{green!10}{Egy:}}    \small Vodafone \<من> iphone \<طيب ما تشغلى خدمة ال> \\
\cline{2-2}

\textbf{mT5} &  \small Vodafone   \<طب ما تشغلي خدمة الأيفون من >      
\\
\cline{2-2}

\textbf{\textbf{AraT5\textsubscript{Tw}}}    &  \small Vodafone   \<طيب ما تشغلي خدمة الأيفون من >   \\

\cdashline{2-2}

\textbf{\textbf{AraT5\textsubscript{MSA}}}    &   \small \<طب ما تشغلي خدمة الأيفون من فودافون >   \\                                         \cdashline{2-2}

\textbf{\textbf{AraT5}}    &  \small{  \<طب ما تشغلي خدمة الأيفون من فودافون > }  \\                                     
\hline 

\textbf{(2) Source } &   \small \colorbox{green!10}{\textbf{Arabizi:} }    3amla eiih enty weih 2a5bar el 5otoba 3la 7eseek?  \\
\cline{2-2}
\textbf{Target} &    \small
\footnotesize{ \colorbox{green!10}{\textbf{Egy:}}   \<عاملة إيه انتي وإيه اخبار الخطوبة على حسك?> 
}\\
           \hline 
\cline{2-2}

\textbf{mT5} &   \small  \<عاملة إيه انتي فيه اخبار الخطوبة على حسابك?>
\\
\cline{2-2}

\textbf{\textbf{AraT5\textsubscript{Tw}}}    &    \small
\<عاملة إيه انتي وإيه اخبار الخطوبة على حسبك?>
\\
\cdashline{2-2}

\textbf{\textbf{AraT5\textsubscript{MSA}}}    &  \small \<عاملة إيه انتي وإيه اخبار الخطوبة على حسابك?>\\                                         \cdashline{2-2}

\textbf{\textbf{AraT5}}    &  \small  \<عاملة إيه انتي وإيه اخبار الخطوبة على بحسبك?>   \\               \hline 

\toprule

\end{tabular}%
}
\caption{\colorbox{green!13}{Transliteration}  examples  extracted frm from Dev data splits.}
    \label{tab:tr_examples}
\end{table*}  
\begin{table*}[t]
\centering
 \renewcommand{\arraystretch}{0.7}
\resizebox{\textwidth}{!}{%
\begin{tabular}{lr}
\toprule 
\rowcolor{red!10}
\textbf{(1) Document:} 
& \begin{tabular}[c]{@{}r@{}}
\<السودان اليوم : اصدر المجلس القومي للمناطق والاسواق الحره برئاسه دكتور مدثر عبدالغني عبدالرحمن وزير الاستثمار>\\
\<قرارا بإلغاء ترخيص عمل شركه قلب العالم الاقتصادية بجزيره مقرسم بولايه البحر الاحمر ووجه القرار الجهات المختصة>\\
\<بضرورة تنفيذه حيث اتخذ المجلس القرار فى اجتماعه الذى انعقد بتاريخ 13 من يونيو الحالى....>\\
\end{tabular}\\   \hline \rowcolor{green!50} 
\textbf{Gold Title:} & \<المجلس القومي الأسواق الحرة.. اصدر قرار بالقاء ترخيص عمل شركة  قلب العالم  >\\
\hline \rowcolor{green!10}
\textbf{  mT5:} 
& \begin{tabular}[c]{@{}r@{}}\<قرار بإلغاء ترخيص عمل شركة قلب العالم الاقتصادية> \end{tabular}\\
\rowcolor{green!10}
\textbf{ AraT5\textsubscript{Tweet}:} 
& \begin{tabular}[c]{@{}r@{}}\<وزير الاستثمار يلغي ترخيص عمل شركة قلب العالم الاقتصادية بجزيره> \end{tabular}\\
\rowcolor{green!10}
\textbf{  AraT5\textsubscript{MSA}:}
& \begin{tabular}[c]{@{}r@{}}\<إلغاء ترخيص شركة قلب العالم الاقتصادية> \end{tabular}\\
\rowcolor{green!10}
\textbf{  AraT5\textsubscript{}:}
& \begin{tabular}[c]{@{}r@{}}\<إلغاء ترخيص عمل شركة قلب العالم الاقتصادية> \end{tabular}\\
\rowcolor{red!10}
\textbf{(2) Document:} 
& \begin{tabular}[c]{@{}r@{}}
\<قال وزير الطاقة التركي فاتح دونميز اليوم الجمعة، إن بلاده حصلت على إعفاء من نحو 25 \% من العقوبات النفطية>\\
\< التي فرضتها الولايات المتحدة على إيران، بما يعادل نحو 3 ملايين طن من النفط سنويا. وقال دونميز في مقابلة مع محطة تلفزيون .....>\\
\end{tabular}\\
\rowcolor{green!50}
 \hline
\textbf{Gold Title:} & \<وزير تركي: إعفاء تركيا بنسبة 25 \% من العقوبات النفطية على إيران> \\
 \hline \rowcolor{green!10}
\textbf{  mT5:} 
& \begin{tabular}[c]{@{}r@{}}\<تركيا تعفي 25 \% من العقوبات النفطية على إيران> \end{tabular}\\
\rowcolor{green!10}
\textbf{ AraT5\textsubscript{Tweet}:} 
& \begin{tabular}[c]{@{}r@{}}\<تركيا تعفي من العقوبات النفطية بنسبة 25\% على إيران> \end{tabular}\\
\rowcolor{green!10}
\textbf{  AraT5\textsubscript{MSA}:}
& \begin{tabular}[c]{@{}r@{}}\<تركيا تحصل على إعفاء من 25 \% من العقوبات النفطية الأمريكية على إيران> \end{tabular}\\
\rowcolor{green!10}
\textbf{  AraT5\textsubscript{}:}
& \begin{tabular}[c]{@{}r@{}}\<تركيا تحصل على إعفاء 25\% من العقوبات الأمريكية على إيران> \end{tabular}\\
\toprule

\end{tabular}%
}
 \caption{Title generation samples  from Dev set using our Models.}
     \label{tab:ngt_examples}
\end{table*}
\label{sec:appendix}
\end{document}